\pdfoutput=1
\documentclass[lettersize,journal]{IEEEtran}
\usepackage{algorithmic}
\usepackage{array}
\usepackage[caption=false,font=normalsize,labelfont=sf,textfont=sf]{subfig}
\usepackage{textcomp}
\usepackage{stfloats}
\usepackage{url}
\usepackage{verbatim}
\usepackage{multirow}
\usepackage{graphicx}
\usepackage{cite}

\usepackage{amsmath,amssymb,amsfonts}
\usepackage{booktabs}
\newcommand{\vect}[1]{\mathbf{#1}}   % vectors/tensors in bold
\newcommand{\mat}[1]{\mathbf{#1}}    % matrices in bold
\newcommand{\R}{\mathbb{R}}

\def\BibTeX{{\rm B\kern-.05em{\sc i\kern-.025em b}\kern-.08em
    T\kern-.1667em\lower.7ex\hbox{E}\kern-.125emX}}
\usepackage{balance}

\begin{document}

\title{PeftCD: Leveraging Vision Foundation Models with Parameter-Efficient Fine-Tuning for Remote Sensing Change Detection}
\author{Sijun Dong, Yuxuan Hu, LiBo Wang, Geng Chen, Xiaoliang Meng\textsuperscript{*}
\thanks{This work was supported by the Major Program (JD) of Hubei Province [grant number: 2023BAA025] and the Key Research and Development Plan of Guangxi Zhuang Autonomous Region [grant number: 2023AB26007].(Corresponding author: Xiaoliang Meng; email: xmeng@whu.edu.cn)}
\thanks{Sijun Dong, Yuxuan Hu and Xiaoliang Meng are with the School of Remote Sensing and Information Engineering, Wuhan University, Wuhan 430079, China (e-mail: \texttt{dyzy41@whu.edu.cn}; \texttt{yuxuanhu@whu.edu.cn}; \texttt{xmeng@whu.edu.cn}).}
\thanks{LiBo Wang is with the School of Remote Sensing and Geomatics Engineering, Nanjing University of Information Science and Technology, Nanjing 210044, China (e-mail: \texttt{rosswanglibo@gmail.com}).}
\thanks{Geng Chen is with the Guangxi Water \& Power Design Institute CO., Ltd.Minzhu road 1-5, Nanning, Guangxi, 530027, China. (e-mail: \texttt{chengeng-hhu@163.com}).}
}

\markboth{Journal of \LaTeX\ Class Files,~Vol.~18, No.~9, September~2025}%
{PeftCD: Leveraging Vision Foundation Models with Parameter-Efficient Fine-Tuning for Remote Sensing Change Detection}

\maketitle

\begin{abstract}
To tackle the prevalence of pseudo changes, the scarcity of labeled samples, and the difficulty of cross-domain generalization in multi-temporal and multi-source remote sensing imagery, we propose \textbf{PeftCD}, a change detection framework built upon \textbf{Vision Foundation Models (VFMs)} with \textbf{Parameter-Efficient Fine-Tuning (PEFT)}. At its core, PeftCD employs a weight-sharing Siamese encoder derived from a VFM, into which LoRA and Adapter modules are seamlessly integrated. This design enables highly efficient task adaptation by training only a minimal set of additional parameters. To fully unlock the potential of VFMs, we investigate two leading backbones: the Segment Anything Model v2 (SAM2), renowned for its strong segmentation priors, and DINOv3, a state-of-the-art self-supervised representation learner. The framework is complemented by a deliberately lightweight decoder, ensuring the focus remains on the powerful feature representations from the backbones. Extensive experiments demonstrate that PeftCD achieves \textbf{state-of-the-art performance} across multiple public datasets, including SYSU-CD (IoU 73.81\%), WHUCD (92.05\%), MSRSCD (64.07\%), MLCD (76.89\%), CDD (97.01\%), S2Looking (52.25\%) and LEVIR-CD (85.62\%), with notably precise boundary delineation and strong suppression of pseudo-changes. In summary, PeftCD presents an optimal balance of accuracy, efficiency, and generalization. It offers a powerful and scalable paradigm for adapting large-scale VFMs to real-world remote sensing change detection applications. The code and pretrained models will be released at https://github.com/dyzy41/PeftCD.
\end{abstract}


\section{Introduction}
Remote sensing change detection (RSCD) is an important research area in remote sensing. It aims to identify surface changes by comparing multi-temporal remote sensing imagery. RSCD is widely used to detect surface changes in industries such as natural resources and water conservancy. In real-world applications, RSCD faces multiple challenges. First, remote sensing data are characterized by multi-temporality, multi-source heterogeneity, and multi-scale variations. Second, existing high-resolution remote sensing datasets are often limited in scale, require costly annotations, and lack sufficient large-scale pretraining samples. In addition, domain shifts caused by heterogeneous sensors and diverse imaging conditions hinder the cross-domain generalization capability of existing models. These issues lead to performance degradation of change detection models under complex scenarios.

In recent years, the rapid development of deep learning has continuously driven the evolution of change detection methods. From early Convolutional Neural Networks (CNNs) to Vision Transformers, and more recently to Vision Foundation Models (VFMs) designed for general-purpose visual tasks, the ability of models to extract features and generalize has been progressively enhanced. VFMs such as SAM~\cite{Kirillov2023SegmentA}, DINO~\cite{Caron2021EmergingPI,simeoni2025dinov3}, and CLIP~\cite{Radford2021LearningTV} are typically pretrained on massive image corpora, exhibiting strong zero-shot and transfer learning abilities across a variety of visual tasks. However, the pretraining data for these models are primarily drawn from natural scene imagery, whereas remote sensing imagery is typically captured from a nadir perspective, featuring complex object compositions and large-scale spatial structures. This significant distribution gap creates performance bottlenecks when directly transferring VFMs to remote sensing change detection.

To address these challenges, it is crucial to both leverage the representational advantages of VFMs and design adaptation strategies tailored for remote sensing tasks. While full-parameter fine-tuning can yield strong task adaptation, it incurs excessive computational and storage costs, limiting deployment in resource-constrained scenarios. Consequently, Parameter-Efficient Fine-Tuning (PEFT) methods have attracted increasing attention in recent years. By freezing most pretrained parameters and only updating a small number of newly introduced modules, PEFT enables efficient task adaptation with performance comparable to full fine-tuning, while significantly reducing training cost.

Motivated by this, we propose a novel change detection framework, termed \textbf{Parameter-Efficient Fine-Tuning Change Detection (PeftCD)}, which builds upon VFM fine-tuning. We select two representative VFMs as backbones: the Segment Anything Model v2 (SAM2) with strong segmentation priors, and DINOv3, a self-supervised representation learner. On top of these, we integrate two canonical PEFT methods, namely LoRA~\cite{LORA} and Adapter~\cite{adapter}, to enable efficient adaptation to remote sensing scenarios. 

Beyond adaptation, we further note that ViT-style backbones such as DINOv3 operate at a fixed spatial resolution after patch embedding and lack an FPN-like pyramid, which may result in blurry boundaries and sensitivity to pseudo-changes when used directly for pixel-level RSCD. To unlock their potential, we design a Multi-layer Fusion and Context Enhancement (MFCE) Decoder for the DINOv3 instantiation (Fig.~\ref{fig:peftcd_dino3cd}). The MFCE decoder integrates same-scale features from multiple Transformer layers through deep attention fusion, and enlarges receptive fields via an ASPP-style context module. In this way, it complements DINOv3’s strong semantic representations without introducing heavy heads. In contrast, the SAM2 instantiation (SAM2CD; Fig.~\ref{fig:peft_sam2cd}) adopts an FPN together with a lightweight ResBlock-based decoder to exploit its segmentation priors.

In summary, the main contributions of this work are as follows:
\begin{itemize}
    \item We propose PeftCD, a novel framework that synergizes leading VFMs (SAM2, DINOv3) with canonical PEFT methods (LoRA, Adapter) to achieve highly efficient adaptation for remote sensing change detection.
    \item To address the limitations of single-scale features in DINOv3, we design a Multi-Layer Fusion and Context Enhancement (MFCE) Decoder that enhances feature representation through same-scale multi-layer fusion and contextual information augmentation.
    \item Extensive experiments on seven public datasets demonstrate that PeftCD achieves state-of-the-art (SOTA) performance with precise boundary delineation and strong suppression of pseudo-changes, validating the effectiveness of our approach.
\end{itemize}

\begin{figure*}[htb]
  \centering
  \includegraphics[width=\textwidth]{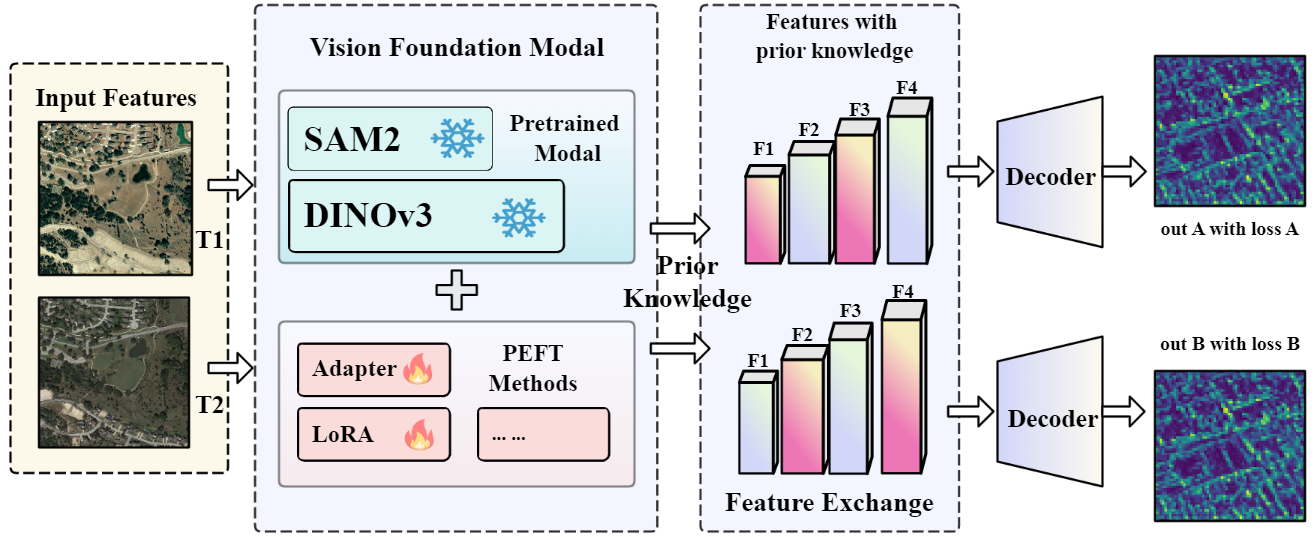}
  \caption{Architecture of PeftCD. Bi-temporal images are encoded by a shared VFM backbone with injected PEFT modules. Features undergo exchange before decoding into a change map.}
  \label{fig:peftcd_framework}
\end{figure*}

\section{Related Work}
\subsection{Overview of Vision Foundation Models}
\subsubsection{Segmentation Foundation Model: SAM2}

Segment Anything Model 2 (SAM2)~\cite{ravi2024sam} is an upgraded version of the original SAM~\cite{Kirillov2023SegmentA}, designed to provide higher-quality and more efficient interactive and automated image segmentation. Trained on the SA-1B dataset, which contains over 1.1 billion segmentation masks, SAM2 has developed a deep understanding of object structures, boundaries, and hierarchical relations. Unlike traditional segmentation models, the core strength of SAM2 lies in its promptable segmentation capability, which allows the model to segment any object in an image based on diverse prompts such as points, bounding boxes, masks, or text. Even in the absence of prompts, the model can automatically generate segmentation results for all objects in a scene.

In change detection tasks, segmentation essentially requires precise delineation of object boundaries to identify newly emerged, disappeared, or morphologically altered objects. With large-scale training, SAM2 exhibits superior boundary-capturing capability, which is highly consistent with the core demands of change detection. In particular, for fine-grained variations in buildings, roads, or water bodies, SAM2 serves as a strong prior, providing high-quality object-level spatial cues and thereby improving overall segmentation accuracy.

\subsubsection{Self-Supervised Representation Learning Model: DINOv3}

DINOv3~\cite{simeoni2025dinov3} represents the latest advancement in self-supervised vision Transformers, marking the frontier of large-scale learning from unlabeled data. Unlike supervised learning, which relies on costly human annotations, self-supervised learning extracts supervisory signals directly from data itself (e.g., predicting masked patches of an image), forcing the model to learn more essential and robust representations in the absence of labels. DINOv3 adopts advanced paradigms such as iBOT~\cite{Zhou2021iBOTIB} and is trained on approximately 17 billion unlabeled images, yielding highly generalizable semantic features across diverse visual tasks.

These features exhibit strong semantic discriminability and are less sensitive to variations in illumination, viewpoint, and color, thereby more stably reflecting structural and semantic properties of objects. In remote sensing change detection, such robustness is particularly critical. Seasonal shifts, lighting changes, and climate variations often introduce a large number of pseudo-changes, which traditional feature-based models may easily misclassify as genuine changes. In contrast, the self-supervised representations learned by DINOv3 effectively suppress these appearance-induced disturbances, enabling more accurate discrimination between true land-cover changes and environmental noise, thus providing stable and reliable feature support for change detection.

\begin{figure*}[htb]
  \centering
  \includegraphics[width=\textwidth]{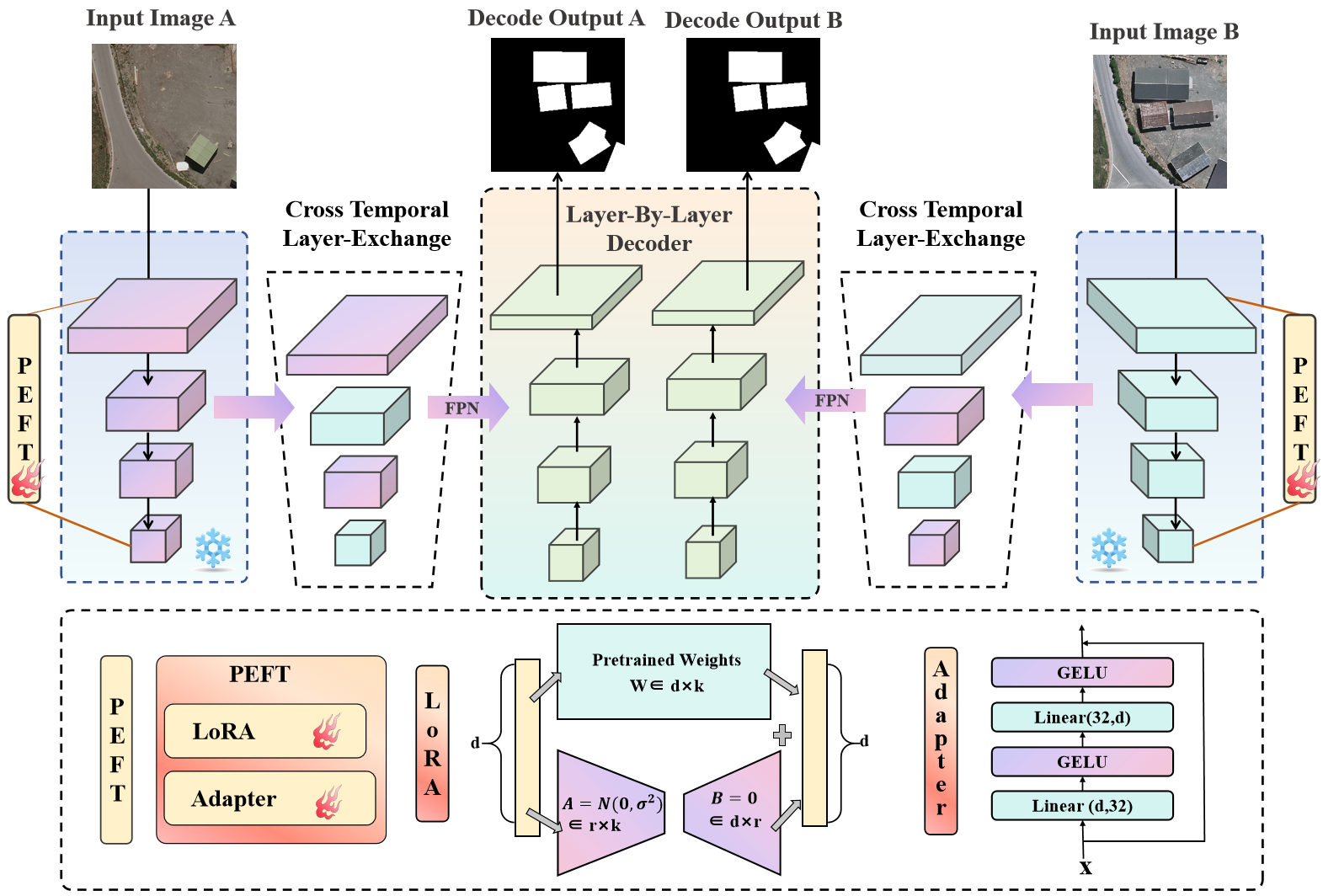}
  \caption{SAM2CD: PeftCD with SAM2 Backbone. Features are exchanged between temporal streams and fused through FPN before dual-branch decoding with shared weights.}
  \label{fig:peft_sam2cd}
\end{figure*}

\subsection{Parameter-Efficient Fine-Tuning Methods}

To effectively adapt foundation models to remote sensing change detection tasks, this work introduces two representative parameter-efficient fine-tuning (PEFT) methods: LoRA~\cite{LORA} and Adapter~\cite{adapter}.

\subsubsection{LoRA (Low-Rank Adaptation)}

LoRA~\cite{LORA} assumes that the weight updates required for downstream task adaptation can be effectively approximated by a low-rank decomposition. Let the original projection weight be $\mat{W}_0\in\mathbb{R}^{d\times k}$. By freezing $\mat{W}_0$, a low-rank increment is introduced as
\begin{equation}
\Delta \mat{W}=\frac{\alpha}{r}\,\mat{B}\mat{A},\qquad
\mat{A}\in\mathbb{R}^{r\times k},\;\mat{B}\in\mathbb{R}^{d\times r},\;r\ll\min(d,k)
\end{equation}
where $r$ denotes the \textit{intrinsic rank}, the bottleneck dimension that controls the number of trainable parameters, and $\alpha$ is a \textit{scaling factor} that adjusts the update magnitude.  
The adapted weight is then given by:
\begin{equation}
\widehat{\mat{W}}=\mat{W}_0+\Delta \mat{W}
\end{equation}
The forward propagation is thus given by
\begin{equation}
\vect{h}=\widehat{\mat{W}}\,\vect{x}
=\mat{W}_0\vect{x}+\frac{\alpha}{r}\,\mat{B}\mat{A}\,\vect{x}
\end{equation}
During training, only $\mat{A}$ and $\mat{B}$ are updated, while $\mat{W}_0$ remains frozen. In this work, LoRA modules are injected into specific linear projections within the Multi-Head Self-Attention (MHSA) of the Transformer. The hyperparameters are set as $r{=}8$ and $\alpha{=}32$. The advantages of LoRA lie in its simplicity and efficiency, as it introduces only a small number of additional parameters while effectively steering the model towards task-specific representations.

\subsubsection{Adapter}

Adapter~\cite{adapter} is one of the earliest proposed PEFT methods. Its core idea is to insert a lightweight ``bottleneck'' structure inside each Transformer block in parallel to the backbone network. An Adapter module typically consists of a down-projection fully connected layer, a nonlinear activation function, and an up-projection fully connected layer. During forward propagation, the input features are first compressed by the down-projection, then passed through the nonlinear activation, and finally restored by the up-projection before being added back to the original features via a residual connection. The flexibility of Adapters lies in their diverse insertion positions, allowing them to enhance task adaptability at specific layers, which makes them particularly suitable for customized fine-tuning. The computation can be formulated as:
\begin{equation}
    h_{\text{Adapter}}(x) = W_{\text{up}} f(W_{\text{down}} x) + x,
    \label{eq:adapter}
\end{equation}
where $x$ denotes the output of a Transformer block, and $f$ is the nonlinear activation function. During fine-tuning, only the internal weights of the Adapter module, $W_{\text{down}}$ and $W_{\text{up}}$, are updated. The advantages of Adapters lie in their structural flexibility and modularity, enabling them to be conveniently integrated into any layer of existing models.

\begin{figure*}[htb]
  \centering
  \includegraphics[width=\textwidth]{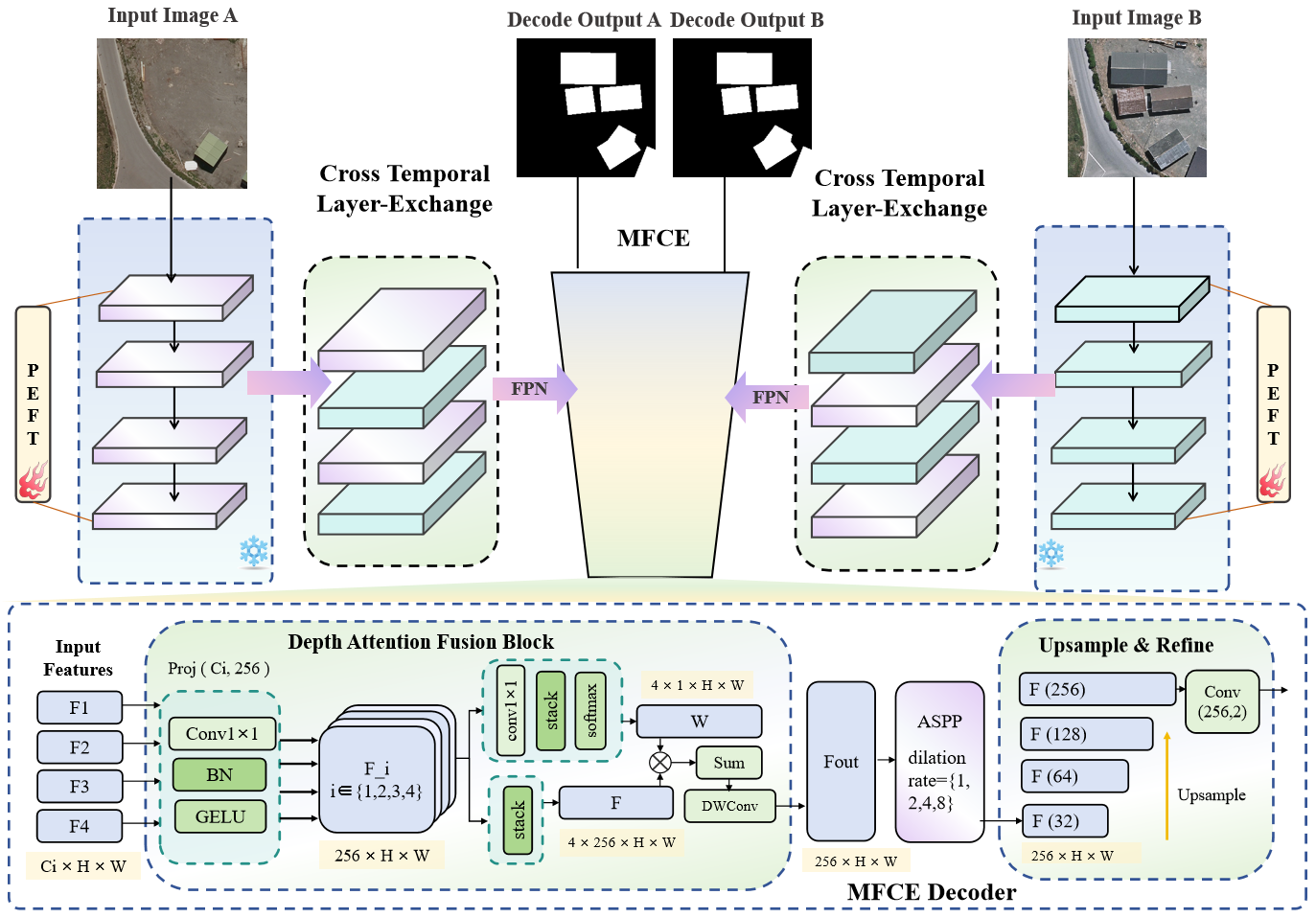} % Ensure the path is correct
  \caption{Architecture of the PeftCD model based on DINOv3 (denoted as DINO3CD).}
  \label{fig:peftcd_dino3cd}
\end{figure*}

\subsection{Change Detection Methods Based on Vision Foundation Models}

Compared with conventional ImageNet-pretrained visual models, Vision Foundation Models (VFMs), such as SAM and DINO, are trained on large-scale datasets, thereby exhibiting stronger generalization capability and segmentation prior knowledge. Consequently, VFMs are generally considered to offer significant prior advantages in remote sensing image analysis. Building on these advantages, recent studies have attempted to incorporate VFMs into change detection in various ways. For instance, Zhang et al.~\cite{sam_feature_extraction} introduced a lightweight feature interaction module attached to the frozen encoder to enhance fine-grained detail representation. Ding et al.~\cite{Ding2023AdaptingSA} combined adapter structures with a semantic branch to achieve sample-efficient modeling of high-resolution remote sensing imagery. Moreover, Saha et al.~\cite{self_sam} integrated unsupervised deep change vector analysis with SAM’s promptable segmentation capability, enabling class-specific change detection with only a small number of samples, thereby substantially reducing reliance on large-scale annotations.

Beyond SAM-based approaches, Zhao et al. combined Dino v2 with adapter-based fine-tuning and proposed Adapter-CD~\cite{Zhao2023AdaptingVT}, which achieved high-accuracy detection on multiple change detection datasets under low-resource constraints. Chen et al.~\cite{Chen2024ChangeDA} introduced Change-DINO, applying the DINO model to object-level change detection. Compared with the SAM family, pixel-level change detection methods based on DINO are relatively scarce. This is because DINO is primarily designed for general-purpose representation learning rather than segmentation-specific tasks. Furthermore, the features output by DINO are typically single-scale high-dimensional semantic embeddings, lacking multi-scale spatial information, which to some extent limits its applicability to pixel-level change detection.

In contrast to the above works, the proposed PeftCD architecture leverages state-of-the-art VFMs (SAM2 and DINOv3) in combination with two canonical parameter-efficient fine-tuning methods (LoRA and Adapter), achieving efficient adaptation for remote sensing change detection. In addition, to address the single-scale feature limitation of DINOv3, we design a decoder that integrates same-scale multi-layer feature fusion with contextual enhancement, thereby fully exploiting its semantic representations and optimizing feature expressiveness in the decoding stage. Overall, PeftCD is designed to be more concise and efficient, while achieving state-of-the-art performance across multiple public datasets, which verifies the strong potential of VFMs combined with PEFT methods for remote sensing change detection.

\begin{table}[htb]
\centering
\caption{PEFT configurations used in PeftCD. LoRA uses dropout=0.1 and targets \texttt{qkv} projections. All encoder weights are frozen and only PEFT modules are trainable. The Adapter wraps each encoder block with a pre-block residual MLP.}
\label{tab:peftcd_cfg}
\begin{tabular}{l l l c c}
\toprule
\textbf{Backbone} & \textbf{Strategy} & \textbf{Injected Layers} & \textbf{Rank/Dim} & \textbf{Scale $\alpha$} \\
\midrule
SAM2   & LoRA    & MHSA (qkv)    & r=8   & 32 \\
SAM2   & Adapter & Pre-block    & dim=32 & -- \\
DINOv3 & LoRA    & MHSA (qkv)     & r=8   & 32 \\
DINOv3 & Adapter & Pre-block     & dim=32 & -- \\
\bottomrule
\end{tabular}
\end{table}

\section{Methods}
\subsection{Overall Framework}

The proposed PeftCD framework is illustrated in Fig.~\ref{fig:peftcd_framework}. Given a pair of bi-temporal remote sensing images $X_t$ and $X_{t'}$, the model first employs a weight-sharing backbone (either SAM2 or DINOv3) to extract feature representations, where \texttt{sam2\_hiera\_large} and \texttt{dinov3\_vitl16} are adopted in our experiments. To achieve parameter-efficient adaptation, we follow the configurations in Table~\ref{tab:peftcd_cfg}. In the LoRA-based strategy, low-rank adapters are injected into the multi-head self-attention \texttt{qkv} projections of every Transformer block, with rank $r{=}8$, scale $\alpha{=}32$, and dropout of $0.1$. In the Adapter-based strategy, each encoder block is augmented with a pre-block residual MLP consisting of a two-layer bottleneck with GELU activation and bottleneck dimension of $32$. In both cases, all backbone weights remain frozen, and only the additional PEFT modules are trainable. After feature extraction, a cross-temporal feature exchange operation is applied between streams to enhance the modeling of change-relevant information. The exchanged features are subsequently decoded using a lightweight prediction head, designed to highlight the representational capacity of the vision foundation model backbones rather than the complexity of the decoder.

The decoding design varies depending on the backbone. For the SAM2-based architecture, we employ a feature pyramid network (FPN) to aggregate multi-scale features and a ResBlock-based progressive upsampling head to produce the final prediction map. In contrast, the DINOv3 backbone produces single-scale, high-dimensional semantic features. To address this, we design a decoder that performs same-scale multi-layer fusion combined with ASPP-based contextual enhancement, followed by progressive upsampling to the original resolution. This design highlights the representational capacity of vision foundation model backbones, while keeping the decoder deliberately lightweight to emphasize adaptation efficiency rather than decoder complexity.

\begin{table}[htb]
\centering
\caption{Quantitative comparison between PeftCD and classic change detection models on the SYSU-CD dataset}
\label{tab:peftcd_sysu}
\begin{tabular}{l c c c c c}
\toprule
\textbf{Model} & \textbf{OA} & \textbf{IoU} & \textbf{F1} & \textbf{Rec} & \textbf{Prec} \\
\midrule
STANet~\cite{chen_spatial-temporal_2020} & 88.24 & 57.22 & 72.79 & 66.71 & 80.08 \\
DSAMNet~\cite{shi_deeply_2022} & -- & 64.18 & 78.18 & 81.86 & 74.81 \\
BIT~\cite{chen_remote_2022} & 90.64 & 66.03 & 79.54 & 77.13 & 82.10 \\
P2V~\cite{lin_transition_2023} & 90.49 & 66.29 & 79.73 & 79.29 & 80.17 \\
HATNet~\cite{Xu2024HybridAT} & 90.92 & 67.00 & 80.24 & 78.23 & 82.36 \\
DARNet~\cite{li_densely_2022} & 91.26 & 68.10 & 81.03 & 79.11 & 83.04 \\
SSANet~\cite{Jiang2022JointVL} & -- & 68.18 & 81.08 & 79.73 & 82.48 \\
CDNeXt~\cite{wei_robust_2024} & 91.99 & 68.57 & 81.36 & 74.10 & 90.20 \\
BASNet~\cite{z_wang_bitemporal_2024} & 91.80 & 69.71 & 82.15 & 80.01 & 84.41 \\
ChangeCLIP~\cite{dong2024changeclip} & 92.08 & 70.53 & 82.72 & 80.41 & 85.16 \\
AGFormer~\cite{Chen2025AGFormerAA} & 92.04 & 70.91 & 82.98 & 82.26 & 83.71 \\
SGSLN~\cite{zhao_exchanging_2023} & -- & 71.05 & 83.07 & 81.45 & 84.76 \\
ChangeMamba~\cite{chen2024changemamba} & 92.30 & 71.10 & 83.11 & 80.31 & 86.11 \\
DEDANet~\cite{Li2025DifferenceEA} & 92.01 & 71.12 & 83.12 & 83.45 & 82.79 \\
DMFANet~\cite{Zhan2025DifferenceAwareMF} & 92.33 & 71.27 & 83.23 & 80.72 & 85.90 \\
UA-BCD~\cite{li_overcoming_2025} & -- & 71.38 & 83.30 & 79.66 & 87.28 \\
DgFA~\cite{f_zhou_dual-granularity_2025}   & 92.42 & 71.86 & 83.63 & 82.08 & 85.24 \\
FTA-Net~\cite{t_zhu_fta-net_2025}   & -- & 72.54 & 84.08 & 82.33 & 85.91 \\
CWmamba~\cite{Liu2025CWmambaLC} & 92.96 & 72.90 & 84.33 & 80.27 & 88.82 \\
SAM-Mamba~\cite{Li2025SAMMambaATC}  & -- & 73.09 & 84.45 & 83.79 & 85.12 \\
DepthCD~\cite{Zhou2025DepthCDDP}   & -- & 73.33 & 84.61 & 83.83 & 85.41 \\
CD-STMamba~\cite{Liu2025CDSTMambaTR} & 93.01 & 73.45 & 84.70 & 82.04 & 87.53 \\

\midrule
PeftCD  & 93.25 & 73.81 & 84.93 & 80.71 & 89.61 \\
\bottomrule
\end{tabular}
\end{table}

\begin{figure*}[!t]
  \centering
  \includegraphics[width=\textwidth]{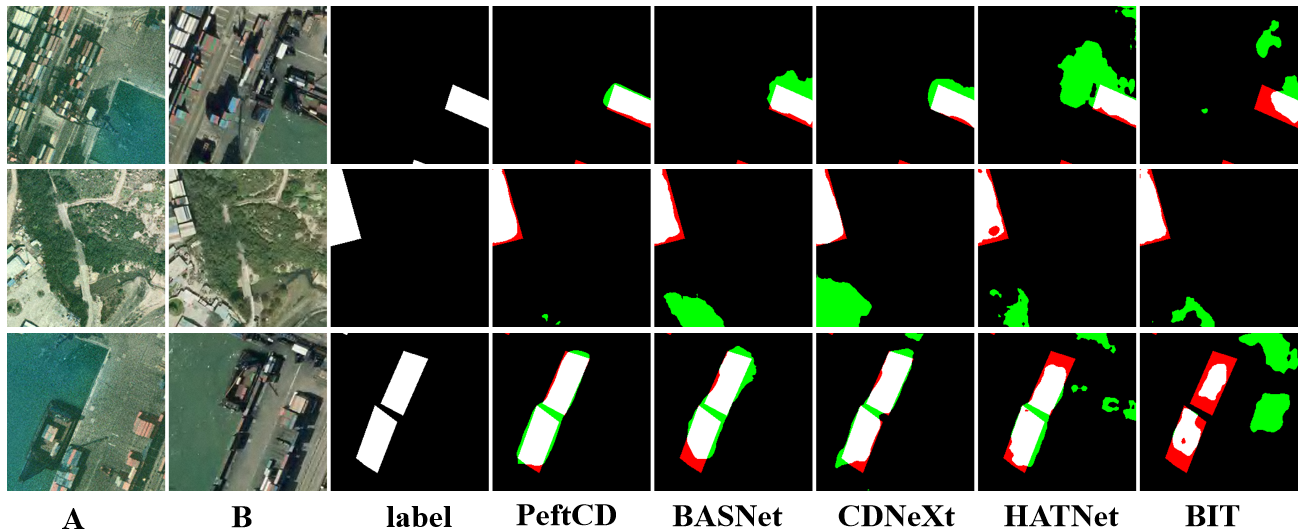}
  \caption{Qualitative comparison between PeftCD and competing methods on the SYSU-CD dataset.}
  \label{fig:peftcd_sysu}
\end{figure*}

\subsection{PeftCD Architecture Based on SAM2}

To effectively transfer the strong segmentation priors of SAM2 to remote sensing change detection, we design the SAM2CD architecture, as illustrated in Fig.~\ref{fig:peft_sam2cd}, which serves as a specific instantiation of the PeftCD framework. The overall structure is divided into the following key components.

\subsubsection{Siamese Encoder with PEFT Injection}
The model adopts a Siamese Encoder structure to process the bi-temporal inputs $T_1$ and $T_2$. The encoder backbone is a weight-sharing SAM2 image encoder, built upon a Vision Transformer (ViT), which is capable of capturing global contextual information of the images. To adapt SAM2 to the change detection task, without altering its large set of pretrained parameters, we intergate PEFT modules (either LoRA or Adapter) into each Transformer block. In each Transformer block, we add LoRA to the MHSA \texttt{qkv} projections with $(r{=}8,\ \alpha{=}32,\ \text{dropout}{=}0.1)$ or, alternatively, wrap the block with a pre-block residual Adapter MLP (bottleneck $32$, GELU). All original SAM2 parameters remain frozen; only the LoRA/Adapter parameters are updated. This design preserves SAM2's strong boundary-aware priors while efficiently steering the representation towards bi-temporal change cues. The encoder outputs a multi-scale feature pyramid for each temporal input. This design allows the model to preserve SAM2’s powerful generic segmentation capability while efficiently learning discriminative features tailored to bi-temporal remote sensing imagery, thereby laying the foundation for subsequent change recognition. The encoder produces two sets of multi-scale feature pyramids, hierarchically representing the corresponding temporal image from low-level textures to high-level semantics.

\subsubsection{Feature Interaction via Layer Exchange}
In the feature interaction stage, we adopt a layer-exchange strategy to encourage the model to learn change-representative features. Specifically, between the two sets of feature pyramids produced by the encoder, features at certain levels (e.g., even layers) are swapped between the two processing streams, while the others (e.g., odd layers) remain unchanged. In this way, two new ``mixed-temporal'' feature pyramids are obtained. These mixed feature pyramids are then fed into a weight-sharing Feature Pyramid Network (FPN) to enhance cross-scale feature fusion, providing richer representations for the decoding stage.

\subsubsection{Siamese Decoder and Supervision}
The decoding stage also adopts a Siamese Decoder structure with shared weights. Each decoder branch receives the mixed-temporal features from the FPN and refines them using ResBlock-based feature extraction, followed by progressive upsampling. Finally, each branch outputs a change probability map. During training, both outputs are supervised with the same ground-truth change mask. During inference, the two probability maps are averaged pixel-wise to obtain a more stable and robust final prediction.

\begin{table}[htb]
\centering
\caption{Quantitative comparison between PeftCD and classic change detection models on the WHUCD dataset}
\label{tab:peftcd_whucd}
\begin{tabular}{l c c c c c}
\toprule
\textbf{Model} & \textbf{OA} & \textbf{IoU} & \textbf{F1} & \textbf{Rec} & \textbf{Prec} \\
\midrule
P2V~\cite{lin_transition_2023} & 99.31 & 85.91 & 92.42 & 90.93 & 93.97 \\
DSIFN~\cite{Zhang2020ADS} & 99.34 & 86.36 & 92.68 & 90.20 & 95.30 \\
MSCANet~\cite{m_liu_cnn-transformer_2022} & 99.36 & 86.65 & 92.85 & 89.98 & 95.90 \\
SGSLN~\cite{zhao_exchanging_2023} & 99.38 & 87.47 & 93.32 & 92.91 & 93.72 \\
BIT~\cite{chen_remote_2022} & 99.43 & 88.22 & 93.74 & 92.00 & 95.56 \\
HCGMNet~\cite{Han2023HCGMNetAH} & 99.52 & 90.10 & 94.79 & 95.31 & 94.27 \\
ChangeCLIP~\cite{dong2024changeclip} & 99.52 & 90.15 & 94.82 & 94.02 & 95.63 \\
CDNeXt~\cite{wei_robust_2024} & 99.54 & 90.35 & 94.93 & 92.56 & 97.43 \\
CGNet~\cite{han_change_2023} & 99.54 & 90.41 & 94.96 & 94.61 & 95.32 \\
EfficientCD~\cite{dong_efficientcd_2024} & 99.55 & 90.71 & 95.13 & 94.19 & 96.08 \\
SAFDNet~\cite{Fu2025BeyondCD} & 99.55 & 90.73 & 95.14 & 94.47 & 95.82 \\
\midrule
PeftCD & 99.62 & 92.05 & 95.86 & 94.95 & 96.79 \\

\bottomrule
\end{tabular}
\end{table}

\begin{figure*}[!t]
  \centering
  \includegraphics[width=\textwidth]{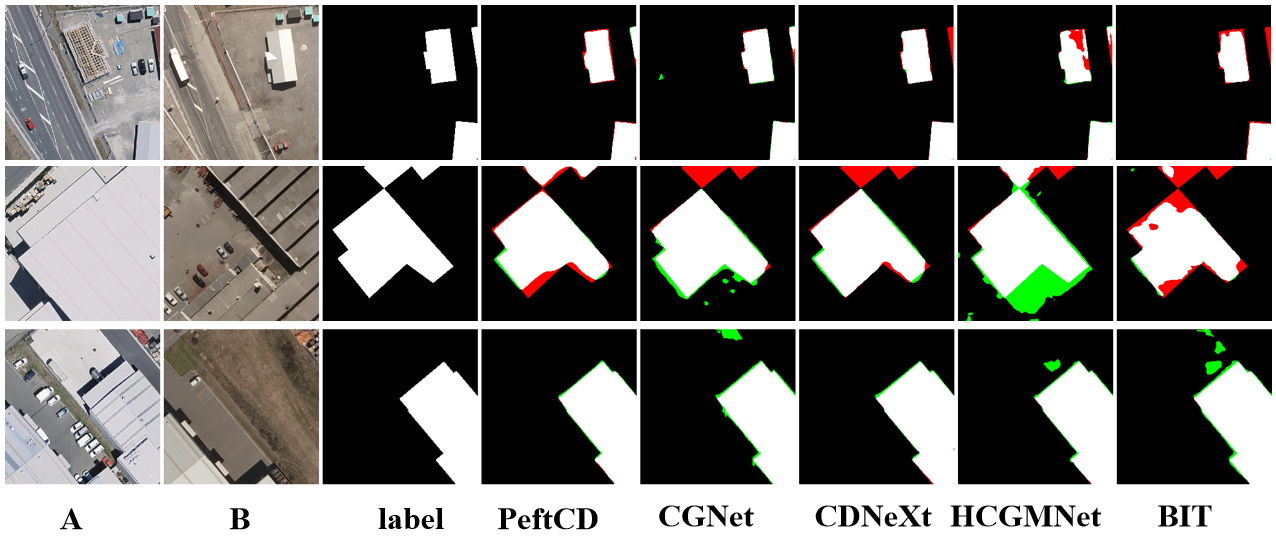}
  \caption{Qualitative comparison between PeftCD and competing methods on the WHUCD dataset.}
  \label{fig:peftcd_whucd}
\end{figure*}

\begin{table}[htb]
\centering
\caption{Quantitative comparison between PeftCD and classic change detection models on the S2Looking dataset}
\label{tab:peftcd_s2looking}
\begin{tabular}{l c c c c c}
\toprule
\textbf{Model} & \textbf{OA} & \textbf{IoU} & \textbf{F1} & \textbf{Rec} & \textbf{Prec} \\
\midrule
HATNet~\cite{Xu2024HybridAT} & - & 47.08 & 64.02 & 60.90 & 67.48 \\
FHD~\cite{pei_feature_2022} & - & 47.33 & 64.25 & 56.71 & 74.09 \\
BIT~\cite{chen_remote_2022} & - & 47.94 & 64.81 & 58.15 & 73.20 \\
SAM-CD~\cite{ding2024adapting} & - & 48.29 & 65.13 & 58.92 & 72.80 \\
DMINet~\cite{feng_change_2023} & - & 48.33 & 65.16 & 62.13 & 68.51 \\
AEGL-Net~\cite{Ying2025AEGLNetAM} & - & 48.36 & 65.19 & 60.69 & 70.05 \\
HFIFNet~\cite{Han2025HFIFNetHF} & 99.22 & 48.54 & 65.35 & 61.04 & 70.33 \\
SGANet~\cite{j_chen_sganet_2025} & - & 48.72 & 66.01 & 57.26 & 77.91 \\
CDNeXt~\cite{wei_robust_2024} & - & 50.05 & 66.71 & 63.08 & 70.78 \\
AGFormer~\cite{Chen2025AGFormerAA} & 99.21 & 50.14 & 66.80 & 65.31 & 68.35 \\
Changer~\cite{fang_changer_2022} & - & 50.47 & 67.08 & 62.04 & 73.01 \\
CWmamba ~\cite{Liu2025CWmambaLC} & 99.24 & 51.43 & 67.93 & 66.19 & 68.38 \\
ConvFormer-CD~\cite{Yang2025ConvFormerCDHC} & 99.21 & 51.51 & 68.00 & 62.81 & 74.12 \\
PeftCD & 99.30 & 52.25 & 68.64 & 63.53 & 74.64 \\
\bottomrule
\end{tabular}
\end{table}

\begin{figure*}[!t]
  \centering
  \includegraphics[width=\textwidth]{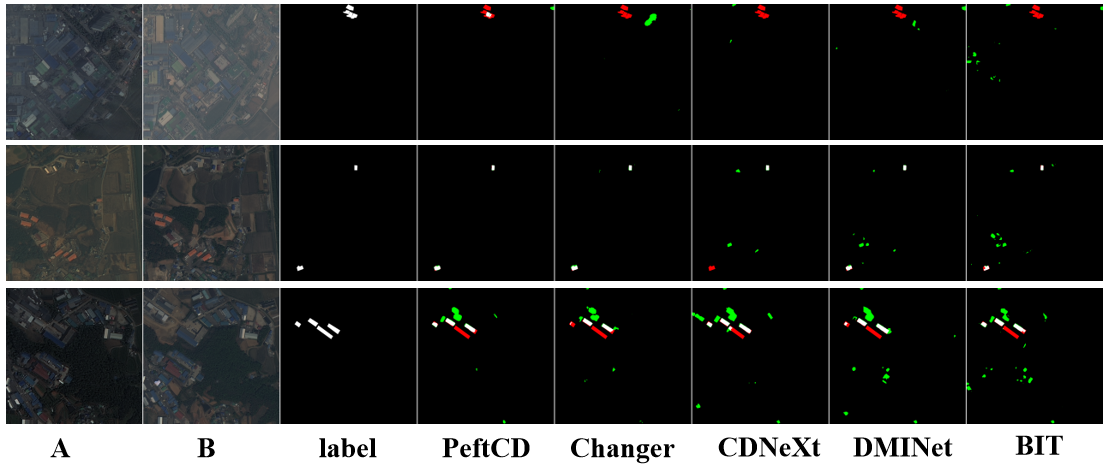}
  \caption{Qualitative comparison between PeftCD and competing methods on the S2Looking dataset.}
  \label{fig:peftcd_s2looking}
\end{figure*}

\subsection{PeftCD Architecture Based on DINOv3}

We also instantiate PeftCD on DINOv3 (Fig.~\ref{fig:peftcd_dino3cd}). Unlike SAM2, DINOv3 is not segmentation-oriented and operates at a fixed spatial resolution after patch embedding, producing same-scale features across layers. Following Table~\ref{tab:peftcd_cfg}, every Transformer block is adapted by LoRA on the MHSA \texttt{qkv} projections ($r{=}8$, $\alpha{=}32$, dropout $0.1$) or by a pre-block residual Adapter MLP (bottleneck $32$, GELU), with the backbone weights frozen and only PEFT parameters trainable. 
After encoding the bi-temporal inputs with a shared-weight encoder, for the \texttt{dinov3\_vitl16} backbone, we select features from layers 5, 11, 17, and 23 to form a multi-level feature sequence, a division analogous to the stages in ResNet~\cite{He2015DeepRL}. Based on this feature sequence, we then perform a feature-layer exchange to enhance temporal interaction. Subsequently, the decoder generates the final change map by applying (i) deep attention fusion across the same-scale multi-layer features, (ii) ASPP-based contextual enhancement, and (iii) progressive upsampling. This design compensates for the lack of a spatial pyramid in ViT-style backbones while preserving DINOv3’s strong semantic representations and cross-domain generalization ability.

\begin{table}[htb]
\centering
\caption{Quantitative comparison between PeftCD and classic models on the MSRSCD dataset}
\label{tab:peftcd_msrscd}
\resizebox{\linewidth}{!}{%
\begin{tabular}{l c c c c c}
\toprule
\textbf{Model} & \textbf{OA} & \textbf{IoU} & \textbf{F1} & \textbf{Rec} & \textbf{Prec} \\
\midrule
STANet~\cite{chen_spatial-temporal_2020} & 91.44 & 54.52 & 70.57 & 69.46 & 71.71 \\
SGSLN~\cite{zhao_exchanging_2023} & 92.52 & 56.28 & 73.36 & 69.73 & 77.39 \\
ChangeFormer~\cite{bandara2022transformer} & 91.86 & 56.96 & 72.58 & 72.94 & 72.22 \\
FCCDN~\cite{Chen2021FCCDNFC} & 92.36 & 57.94 & 73.37 & 71.30 & 75.56 \\
AANet~\cite{Hang2024AANetAA} & 92.17 & 59.23 & 74.40 & 77.03 & 71.94 \\
EATDer~\cite{Ma2024EATDerEA} & 91.47 & 59.29 & 74.44 & 84.17 & 66.73 \\
MSNet~\cite{Liu2025NetworkAD} & 93.03 & 59.80 & 75.74 & 74.73 & 76.79 \\
DMINet~\cite{feng_change_2023} & 93.05 & 61.24 & 75.96 & 74.37 & 77.63 \\
BASNet~\cite{z_wang_bitemporal_2024} & 92.87 & 61.31 & 76.02 & 76.51 & 75.53 \\
CDNeXt~\cite{wei_robust_2024} & 93.24 & 61.59 & 76.23 & 73.42 & 79.27 \\
DFFNet~\cite{Liu2025FullScaleCD} & 93.23 & 61.82 & 77.44 & 78.69 & 76.22 \\
\midrule
PeftCD & 93.83 & 64.07 & 78.10 & 74.19 & 82.44 \\
\bottomrule
\end{tabular}%
}
\end{table}

\begin{figure*}[!t]
  \centering
  \includegraphics[width=\textwidth]{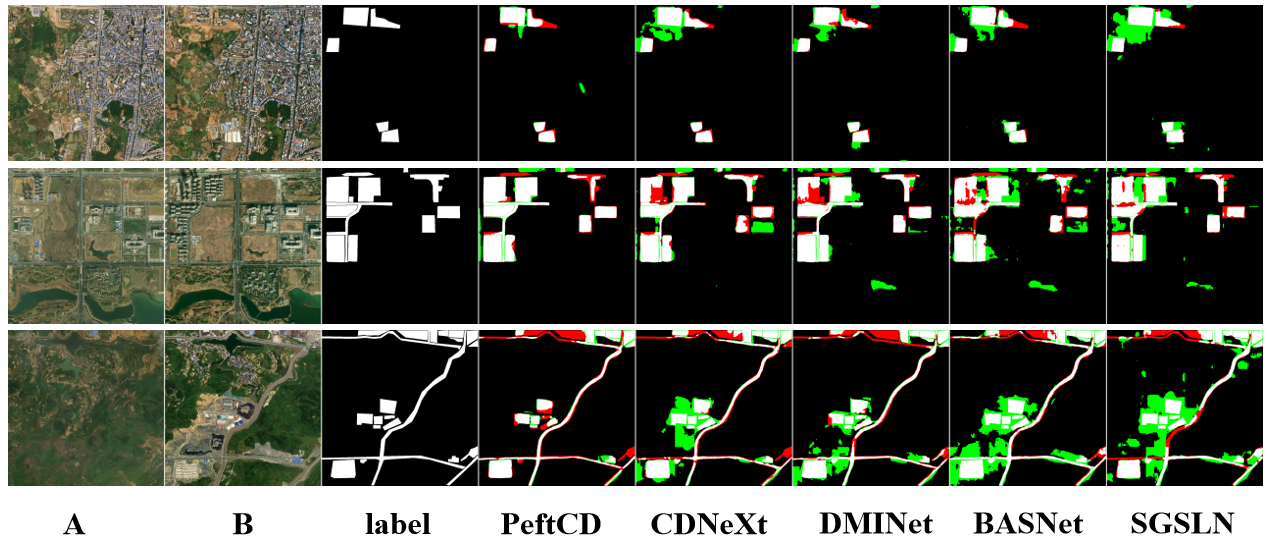}
  \caption{Qualitative comparison between PeftCD and competing methods on the MSRSCD dataset.}
  \label{fig:peftcd_msrscd}
\end{figure*}

\begin{table}[htb]
\centering
\caption{Quantitative comparison between PeftCD and classic change detection models on the MLCD dataset}
\label{tab:peftcd_mlcd}
\resizebox{\linewidth}{!}{
\begin{tabular}{l c c c c c}
\toprule
\textbf{Model} & \textbf{OA} & \textbf{IoU} & \textbf{F1} & \textbf{Rec} & \textbf{Prec} \\
STANet~\cite{chen_spatial-temporal_2020}         & 86.97 & 60.87 & 75.67 & 73.4  & 78.09 \\
ISDANet~\cite{h_ren_interactive_2025}            & 88.3  & 66.52 & 79.89 & 84.2  & 76    \\
EFI-SAM~\cite{Huang2025SAMBasedEF}          & -- & 67.72 & 80.75 & -- & -- \\
DARNet~\cite{li_densely_2022}         & 90.24 & 69.05 & 81.69 & 78.84 & 84.75 \\
BASNet~\cite{z_wang_bitemporal_2024}         & 90.51 & 70.55 & 82.73 & 82.29 & 83.17 \\
BIT~\cite{chen_remote_2022}            & 91.23 & 72.02 & 83.73 & 81.74 & 85.83 \\
DMINet~\cite{feng_change_2023}         & 91.94 & 74.03 & 85.08 & 83.18 & 87.07 \\
\midrule
PeftCD & 92.94 & 76.89 & 86.93 & 85.02 & 88.94 \\
\bottomrule
\end{tabular}}
\end{table}

\begin{figure*}[!t]
  \centering
  \includegraphics[width=\textwidth]{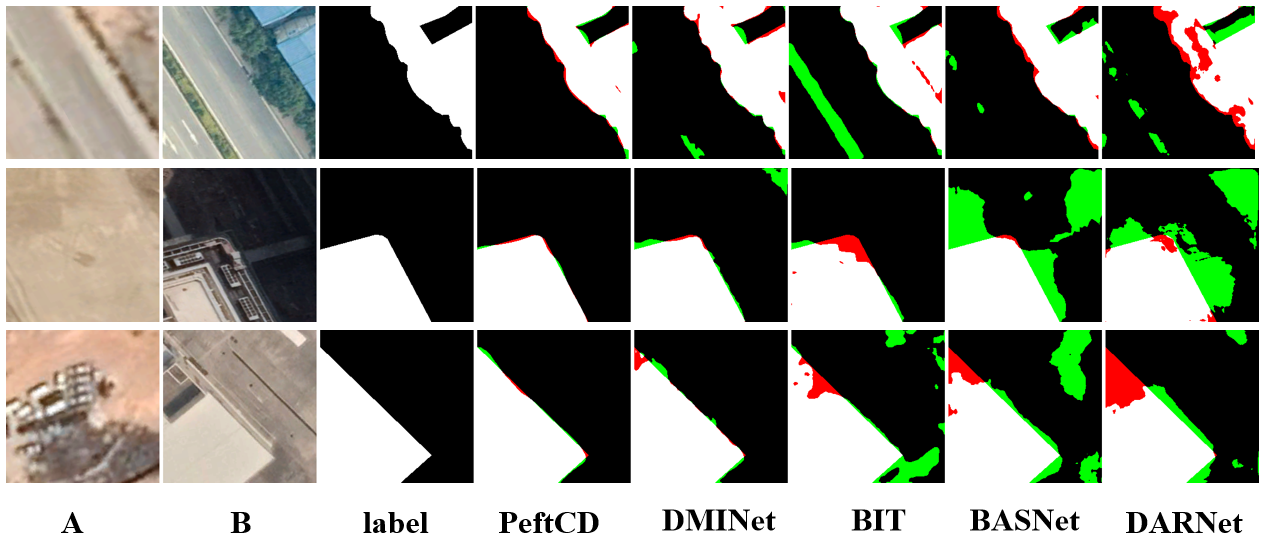}
  \caption{Qualitative comparison between PeftCD and competing methods on the MLCD dataset.}
  \label{fig:peftcd_mlcd}
\end{figure*}

\subsection{Decoder with Same-Scale Multi-Layer Fusion and Contextual Enhancement (MFCE)}

Traditional CNN-based encoder–decoder architectures typically rely on multi-scale feature pyramids (Feature Pyramid Networks, FPNs) generated by the encoder. These pyramids, produced at different downsampling stages, capture high-level semantic information (deep, low-resolution features) and fine-grained spatial details (shallow, high-resolution features), which are crucial for the decoder to progressively restore resolution and achieve accurate segmentation~\cite{lin_feature_2017}.

However, when employing VFMs based on DINO or other Vision Transformers (ViTs) as the encoder backbone, this classical paradigm faces challenges. In the initial patch embedding stage, ViT architectures downsample the input image by a factor of 16 and tokenize it into image patches. All subsequent Transformer blocks operate at this fixed resolution (e.g., $H/16 \times W/16$), performing feature transformation and interaction. Although the output features from different Transformer layers capture hierarchical semantic levels ranging from low-level to high-level, they remain single-scale in spatial resolution and thus do not form a traditional feature pyramid. This architectural characteristic leads to a shortage of multi-scale spatial information during upsampling in the decoder, which is particularly detrimental for recovering object boundaries and fine-grained changes, potentially resulting in blurred edges and missed small objects in the prediction.

To address the absence of feature pyramids in ViT backbones, we design a dedicated decoder. Instead of relying on multi-scale spatial inputs, the decoder builds strong feature representations on single-scale feature maps through two core steps, which are then used as the basis for upsampling. First, to fully leverage the hierarchical features from different ViT layers, we perform deep attention fusion across same-scale multi-layer features. Second, to enhance contextual understanding during decoding, we incorporate an Atrous Spatial Pyramid Pooling (ASPP) module to construct a receptive field pyramid for context enhancement. Consequently, the input to the decoder combines both hierarchical feature integration and rich multi-scale contextual awareness. Finally, a progressive upsampling path restores spatial resolution step by step to generate the final change detection map.

\begin{table}[htb]
\centering
\caption{Quantitative comparison between PeftCD and classic change detection models on the CDD dataset}
\label{tab:peftcd_cdd}
\resizebox{\linewidth}{!}{
\begin{tabular}{l c c c c c}
\toprule
\textbf{Model} & \textbf{OA} & \textbf{IoU} & \textbf{F1} & \textbf{Rec} & \textbf{Prec} \\
\midrule
BASNet~\cite{z_wang_bitemporal_2024} & 99.18 & 93.29 & 96.53 & 96.30 & 96.76 \\  
UA-BCD~\cite{li_overcoming_2025}     & --    & 93.49 & 96.64 & 96.90 & 96.38 \\
RCDT~\cite{lu_cross_2024}                  & --    & 93.79 & 96.80 & 96.97 & 96.63 \\
CDMamba~\cite{zhang_cdmamba_2025}    & 99.26 & 93.93 & 96.87 & 96.84 & 96.90 \\
SFFCE-CD~\cite{y_xing_sffce-cd_2025}   & 99.29 & 94.42 & 97.13 & 96.94 &  97.32 \\
ELGCNet~\cite{m_noman_elgc-net_2024}   & 99.33  &  94.50   & 91.17  & --   & -- \\   
DgFA~\cite{f_zhou_dual-granularity_2025} & 99.41 & 95.12 & 97.50 & 97.60 & 97.40 \\
GASNet~\cite{zhang_global-aware_2023}  & 99.41  & 95.34  & 97.61  & 98.06  & 97.17 \\
DEDANet~\cite{Li2025DifferenceEA} & 99.40 & 95.46 & 97.67 & 97.14 & 98.22 \\
DSCRNet ~\cite{Zhang2025ADS} & -- & 95.46 & 97.68 & 97.85 & 97.51 \\
CD-STMamba~\cite{Liu2025CDSTMambaTR} & 99.41 & 95.56 & 97.73 & 97.45 & 98.01 \\
MLDFNet~\cite{d_sidekejiang_mldfnet_2025}  & -- & 95.78 & 97.84 & 97.97 & 97.72 \\
ScratchFormer~\cite{Noman2023RemoteSC}   & 99.50 & 95.85  & 97.88 & -- & -- \\ 
FTransDF-Net~\cite{li_dual_2025}   & -- & 95.85 &  97.88 & 97.63 & 98.13 \\
DSFI-CD~\cite{x_li_dsfi-cd_2025}  & 99.52 & 96.10 &  98.01 & 98.34 & 97.68 \\
HASNet~\cite{c_tao_hasnet_2025}        & 99.56 & 96.49 & 98.21 & 98.12 & 98.32 \\
\midrule
PeftCD & 99.63 & 97.01 & 98.48 & 98.30 & 98.67 \\
\bottomrule
\end{tabular}}
\end{table}

\begin{figure*}[!t]
  \centering
  \includegraphics[width=\textwidth]{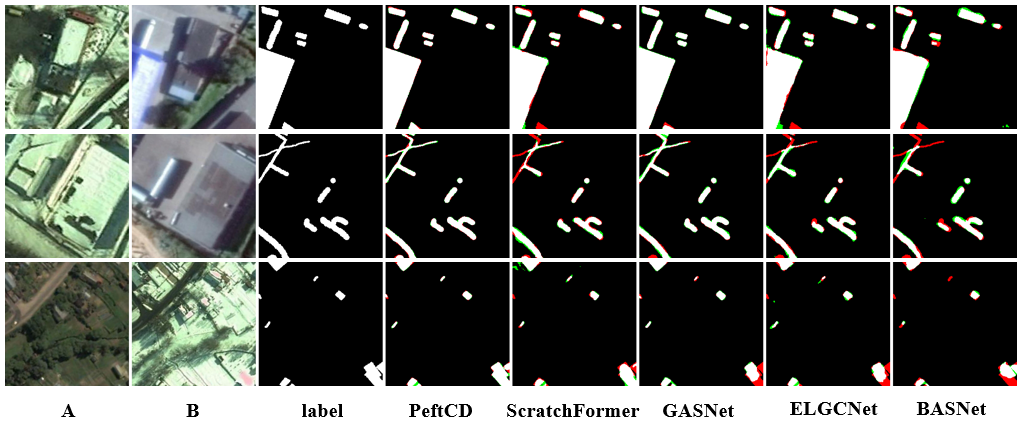}
  \caption{Qualitative comparison between PeftCD and competing methods on the CDD dataset.}
  \label{fig:peftcd_cdd}
\end{figure*}

\subsubsection{Deep Attention Fusion of Same-Scale Multi-Layer Features}
Although ViT backbones maintain a single-scale spatial resolution, their outputs at different layers exhibit varying levels of semantic abstraction: shallow layers focus on local textures and boundaries, while deeper layers encode global semantics and category-level information. Effectively fusing these ``same-scale but hierarchically different'' features is thus critical for enhancing discriminative power.

Let the feature from the $i$-th encoder layer be $\{\vect{F}_i\}_{i=1}^{N}$, where $\vect{F}_i\in\R^{B\times C_i\times H\times W}$, $B$ is the batch size, $C_i$ is the channel dimension, and $H$, $W$ are the spatial dimensions. For cross-layer fusion, we first use $1\times1$ convolutions to project each feature into a unified dimension $C_{\mathrm{mid}}$, yielding $\vect{X}_i\in\R^{B\times C_{\mathrm{mid}}\times H\times W}$.

On this basis, we design position-adaptive attention weights for each spatial location $(h,w)$ to dynamically determine which semantic level is most informative. Specifically, a lightweight $1\times1$ convolution generates a score map $S_i$ for each $\vect{X}_i$, followed by softmax normalization across layers:
\begin{equation}
A_i(h,w)=\frac{\exp\big(S_i(h,w)\big)}{\sum_{j=1}^{N}\exp\big(S_j(h,w)\big)},\qquad
\sum_{i=1}^{N}A_i(h,w)=1 .
\end{equation}

The fused feature at each spatial position is then obtained as a weighted sum of all layers:
\begin{equation}
\vect{X}_{\mathrm{fused}}(h,w)=\sum_{i=1}^{N} A_i(h,w)\,\vect{X}_i(h,w).
\end{equation}

The advantage of this deep attention fusion mechanism is that it does not assign fixed global weights to layers; instead, each spatial location adaptively selects the most discriminative level of information. For instance, at boundary pixels of change regions, shallow texture and edge features may be emphasized, whereas in the interior of large-scale objects, deeper semantic features dominate. The fused representation $\vect{X}_{\mathrm{fused}}$ thus retains local details while incorporating global semantics, providing a solid foundation for subsequent ASPP-based context enhancement and progressive decoding.

\subsubsection{Contextual Enhancement and Progressive Upsampling}
After feature fusion, we employ an Atrous Spatial Pyramid Pooling (ASPP) module to further enhance contextual awareness. ASPP applies parallel depthwise separable dilated convolutions with multiple dilation rates on the fused feature map, thereby capturing multi-scale contextual information without reducing spatial resolution. This effectively constructs a receptive field pyramid, compensating for the lack of spatial scale diversity in the ViT backbone and providing robust semantic context for subsequent upsampling.

Finally, the context-enhanced feature map (still at $1/16$ resolution) is passed through a progressive upsampling path. This path consists of three consecutive stages, each combining bilinear interpolation with depthwise separable convolutions, gradually restoring the resolution from $1/16$ to $1/8$, $1/4$, and eventually to the original input size. The depthwise separable convolution modules refine and smooth the features after each upsampling stage, enriching fine details. The prediction head then produces pixel-level change detection results.

\begin{table}[htb]
\centering
\caption{Quantitative comparison between PeftCD and classic change detection models on the LEVIR-CD dataset}
\label{tab:peftcd_levir}
\resizebox{\linewidth}{!}{
\begin{tabular}{l c c c c c}
\toprule
\textbf{Model} & \textbf{OA} & \textbf{IoU} & \textbf{F1} & \textbf{Rec} & \textbf{Prec} \\
\midrule
    STANet~\cite{chen_spatial-temporal_2020}   & 99.02 & 81.85 & 90.02 & 87.13 & 93.10 \\
    MSCANet~\cite{m_liu_cnn-transformer_2022}          &  99.03 &  81.91 &  90.06 &  86.38 &  94.06  \\
    MFATNet~\cite{Mao2022MFATNetMF}          &  99.03 &  82.42 &  90.36 &  88.93 &  91.85  \\
    ChangeFormer~\cite{bandara2022transformer}     &  99.04 &  82.66 &  90.50 &  90.18 &  90.83  \\
    P2V~\cite{lin_transition_2023}              &  99.04 &  83.00 &  90.71 &  91.78 &  89.67  \\
    CDMamba~\cite{zhang_cdmamba_2025}          &  99.06 &  83.07 &  90.75 &  90.08 &  91.43  \\
    AMTNet~\cite{Liu2023AnAM}           &   --   &  83.08 &  90.76 &  89.71 &  91.82  \\
    B2CNet~\cite{Zhang2024B2CNetAP} & 99.10 & 83.72 & 91.14 & 91.01 & 91.27 \\
    DgFA~\cite{f_zhou_dual-granularity_2025}   & 99.11 & 83.93 & 91.26 & 90.84 & 91.69 \\
    HATNet~\cite{Xu2024HybridAT} & 99.13 & 84.12 & 91.38 & 90.15 & 92.64 \\
    DMATNet~\cite{Song2022RemoteSI}          &  98.25 &  84.13 &  90.75 &  89.98 &  91.56  \\
    DEDANet~\cite{Li2025DifferenceEA}          &  99.14 &  84.18 &  91.41 &  90.06 &  92.80  \\
    STransUNet~\cite{Yuan2022STransUNetAS}       &  99.13 &  84.19 &  91.41 &  90.55 &  92.30  \\
    LCCDMamba~\cite{Huang2025LCCDMambaVS}       &  99.16 &  84.64 &  91.68 &  92.42 &  90.96  \\
    AGFormer~\cite{Chen2025AGFormerAA}         &  99.17 &  84.81 &  91.78 &  90.84 &  92.74  \\
    DSCRNet~\cite{Zhang2025ADS}         &  -- &  85.15 &  91.98 &  92.43 &  91.53  \\
    HCGMNet~\cite{Han2023HCGMNetAH}          &  99.18 &  85.26 &  92.04 &  92.81 &  91.29  \\
    Changer~\cite{fang_changer_2022}        &   --   &  85.29 &  92.06 &  90.56 &  93.61  \\
    DMFANet~\cite{Zhan2025DifferenceAwareMF}         &  99.20 &  85.33 &  92.09 &  90.93 &  93.27  \\
    CGNet~\cite{han_change_2023}            &  99.20 &  85.40 &  92.13 &  91.93 &  92.32  \\
    EfficientCD~\cite{dong_efficientcd_2024} &  99.22 &  85.55 &  92.21 &  91.22 &  93.23  \\
    FTA-Net~\cite{t_zhu_fta-net_2025}   & -- & 85.58 & 92.23 & 92.68 & 91.79 \\
\midrule
PeftCD  & 99.22 & 85.62 & 92.25 & 91.44 & 93.08 \\
\bottomrule
\end{tabular}}
\end{table}

\begin{figure*}[!t]
  \centering
  \includegraphics[width=\textwidth]{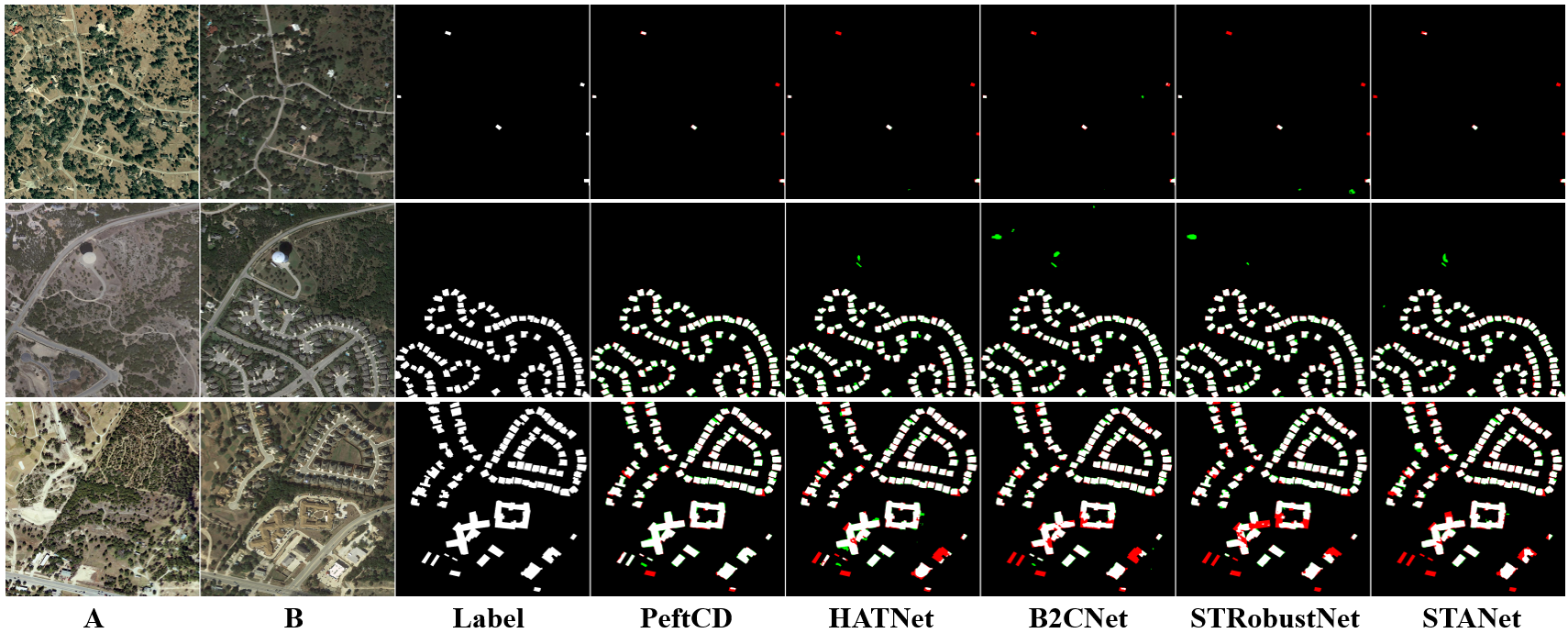}
  \caption{Qualitative comparison between PeftCD and competing methods on the LEVIR-CD dataset.}
  \label{fig:peftcd_levir}
\end{figure*}

\section{Experiment}
\subsection{Datasets}
To comprehensively assess the effectiveness and generalizability of the proposed method, we evaluate it on seven representative remote sensing change-detection datasets: SYSU-CD~\cite{shi_deeply_2022}, WHUCD~\cite{Ji2019FullyCN}, MSRSCD~\cite{Liu2025NetworkAD}, CDD~\cite{Lebedev2018CHANGEDI}, S2Looking~\cite{Shen2021S2LookingAS}, LEVIR-CD~\cite{chen_spatial-temporal_2020}, and MLCD~\cite{Huang2025SAMBasedEF}. These datasets span diverse spatial resolutions, geographic regions, and change types, providing a broad basis for evaluating model robustness and cross-scenario adaptability. A brief summary is given below.

\textbf{SYSU-CD}~\cite{shi_deeply_2022} contains 20{,}000 pairs of high-resolution aerial images ($256\!\times\!256$) over Hong Kong (2007–2014), covering changes such as new buildings, suburban expansion, offshore construction, vegetation dynamics, and road extensions. Following the common split, 12{,}000/4{,}000/4{,}000 pairs are used for training/validation/testing.

\textbf{WHUCD}~\cite{Ji2019FullyCN} consists of two large-format aerial images (about $32{,}507\!\times\!15{,}354$ at 0.3\,m/pixel) designed for building change detection. For fair comparison, the images are tiled into non-overlapping $256\!\times\!256$ patches, yielding approximately 6{,}096/762/762 samples for train/val/test.

\textbf{MSRSCD}~\cite{Liu2025NetworkAD} provides 842 pairs of $1024\!\times\!1024$ bi-temporal images with multiple target categories (e.g., buildings, roads, industrial sites) and a balanced object-size distribution, making it suitable for evaluating multi-scale detection capability.

\textbf{CDD}~\cite{Lebedev2018CHANGEDI} comprises satellite image pairs capturing seasonal variations over the same regions, with ground sampling distances from 0.03\,m to 1\,m. A typical split adopts 10{,}000/3{,}000/3{,}000 samples for train/val/test, targeting the discrimination of true temporal changes from appearance shifts.

\textbf{S2Looking}~\cite{Shen2021S2LookingAS} focuses on building changes under side-looking satellite imagery, which introduces pronounced viewpoint distortions and illumination variations. It contains 5{,}000 bi-temporal pairs (around $1024\!\times\!1024$, 0.5–0.8\,m) with 65{,}920 annotated change instances across rural regions worldwide, posing greater challenges than nadir-view benchmarks.

\textbf{LEVIR-CD}~\cite{chen_spatial-temporal_2020} is a widely used building-change benchmark with 637 pairs of high-resolution Google Earth images ($1024\!\times\!1024$, $\sim$0.5\,m) from U.S. urban areas and 31{,}333 annotated building-change instances. The dataset is commonly split into 445/128/64 pairs for train/val/test.

\textbf{MLCD}~\cite{Huang2025SAMBasedEF} (Macao Land Change Detection) targets land reclamation and vegetation dynamics in Macao from 2008 to 2023 using Google Earth Engine imagery. It contains 10{,}000 labeled pairs of $256\!\times\!256$ images with 0.5–2\,m resolution, providing a new benchmark for coastal geomorphology and land-use evolution studies.

\subsection{Implementation Details}
We implement the proposed method in PyTorch with extensions from the Pytorch Lightning ecosystem. Experiments are conducted on Ubuntu with four NVIDIA GeForce RTX 4090 GPUs, enabling efficient distributed training. All models including baselines and PeftCD are trained under the same settings with a batch size of 16. Standard geometric data augmentations are applied to improve generalization. We use the AdamW optimizer with an initial learning rate of 0.0003 and a weight decay of 0.01. Throughout training, we monitor the change-class Intersection over Union (IoU) on the validation set and select the checkpoint with the best validation performance for testing. Additionally, for the LEVIR-CD, S2Looking, and MSRSCD datasets, we adopt random $256\times256$ cropping during training and employ sliding-window inference at test time. PeftCD is optimized with cross-entropy loss, and we apply a 3{,}000-step warm-up.

\subsection{Evaluation Metrics}
To thoroughly evaluate the performance of the proposed method, we adopt several widely used metrics, including Overall Accuracy (OA), Intersection over Union (IoU), F1 score, Recall (Rec), and Precision (Prec). A higher precision indicates fewer false positives (FP), while a higher recall reflects fewer false negatives (FN). These metrics collectively assess the detection capability of a model from different perspectives, and higher values consistently indicate better performance. Their formulations are given as follows:

\begin{equation}
OA = \frac{TP + TN}{TP + TN + FN + FP}
\end{equation}

\begin{equation}
IoU = \frac{TP}{TP + FN + FP}
\end{equation}

\begin{equation}
Rec = \frac{TP}{TP + FN}
\end{equation}

\begin{equation}
Prec = \frac{TP}{TP + FP}
\end{equation}

\begin{equation}
F1 = \frac{2 \cdot Prec \cdot Rec}{Prec + Rec}
\end{equation}

where $TP$, $TN$, $FP$, and $FN$ denote the numbers of true positives, true negatives, false positives, and false negatives, respectively.

\subsection{Compared Methods}
We benchmark PeftCD against a representative set of baselines spanning three methodological paradigms: classic CNN/Transformer methods (e.g., STANet~\cite{chen_spatial-temporal_2020}, CGNet~\cite{han_change_2023}, BIT~\cite{chen_remote_2022}, ChangeFormer~\cite{bandara2022transformer}, EfficientCD~\cite{dong_efficientcd_2024}), Mamba-family state-space models (e.g., ChangeMamba~\cite{chen2024changemamba}, CD-STMamba~\cite{Liu2025CDSTMambaTR}, CWmamba~\cite{Liu2025CWmambaLC}), and SAM-based approaches that inject segmentation priors (e.g., SAM-CD~\cite{ding2024adapting}, EFI-SAM~\cite{Huang2025SAMBasedEF}, SAM-Mamba~\cite{Li2025SAMMambaATC}). This selection balances conventional architectures, modern state-space sequence modeling, and VFM-guided variants, providing a concise yet representative benchmark for comparison.

\subsection{Results and Analysis}

We conduct extensive quantitative experiments on seven challenging datasets, SYSU-CD, WHUCD, MSRSCD, CDD, S2Looking, LEVIR-CD, and MLCD, to comprehensively evaluate the effectiveness of the proposed PeftCD framework, and compare it against representative state-of-the-art change detection methods. As summarized in Tables~\ref{tab:peftcd_sysu},~\ref{tab:peftcd_whucd},~\ref{tab:peftcd_s2looking},~\ref{tab:peftcd_msrscd},~\ref{tab:peftcd_mlcd},~\ref{tab:peftcd_cdd} and~\ref{tab:peftcd_levir}, the results provide strong evidence of the superiority of our approach. Across all datasets, PeftCD equipped with PEFT strategies clearly outperforms conventional CNN- and Transformer-based baselines and achieves performance comparable to, or surpassing, the current SOTA methods.

More concretely, on SYSU-CD, WHUCD, MSRSCD, CDD, S2Looking, and MLCD, PeftCD attains IoU scores of 73.81\%, 92.05\%, 64.07\%, 97.01\%, 52.25\%, and 76.89\%, respectively, outperforming all competing approaches. On SYSU-CD, PeftCD improves IoU by 2.71 percentage points to 73.81\% and achieves an F1 score of 84.93\%, significantly higher than the runner-up ChangeMamba (83.11\%). On WHUCD, PeftCD reaches 92.05\% IoU, exceeding EfficientCD by 1.34 percentage points, and obtains a Precision of 96.79\%, indicating strong suppression of false positives. On S2Looking, PeftCD achieves an IoU of 52.25\%, outperforming Conv-former (51.51\%). These results demonstrate that PeftCD excels in complex change scenarios and effectively captures diverse change patterns.

To further illustrate the advantages of PeftCD in practical detection, we provide qualitative comparisons in Figures~\ref{fig:peftcd_sysu},~\ref{fig:peftcd_whucd},~\ref{fig:peftcd_s2looking},~\ref{fig:peftcd_msrscd},~\ref{fig:peftcd_mlcd},~\ref{fig:peftcd_cdd} and~\ref{fig:peftcd_levir}. In these visualizations, white pixels denote correctly detected change regions (TP), black denotes correctly detected non-change regions (TN), green indicates false positives (FP), and red indicates false negatives (FN).

From the qualitative results, PeftCD consistently exhibits several advantages over competing methods. First, it more completely identifies change regions, markedly reducing missed detections (red), especially for small objects and irregular shapes (e.g., scattered buildings in SYSU-CD and small detached buildings in WHUCD). Second, it delivers more precise and smoother boundary delineation of change regions, effectively mitigating the boundary blurring or jagged artifacts commonly observed in other models. Most importantly, PeftCD shows strong robustness in suppressing background noise and pseudo-changes, yielding substantially fewer false positives (green). We attribute these benefits to the strong semantic understanding and generalization capability of vision foundation models, which enable better discrimination between true land-cover changes and appearance variations induced by illumination, seasonal shifts, or other environmental factors. In summary, both qualitative and quantitative analyses consistently confirm that PeftCD achieves leading performance in terms of detection accuracy, completeness, and robustness.

\begin{table*}[!htb]
\centering
\caption{Ablation results of different backbone models and PEFT methods (IoU \%).}
\label{tab:peftcd_ablation}
\begin{tabular}{l l c c c c c c c} % ← 9 列
\toprule
\textbf{Backbone} & \textbf{Fine-tuning Method} & \textbf{SYSU-CD} & \textbf{WHUCD} & \textbf{S2Looking} & \textbf{MSRSCD} & \textbf{MLCD} & \textbf{LEVIR-CD} & \textbf{CDD} \\
\midrule
\multirow{3}{*}{SAM2}
& Frozen Encoder            & 69.45 & 87.83 & 47.98 & 60.42 & 73.49 & 83.58 & 95.95 \\
& Adapter                   & 71.91 & 91.81 & 49.97 & 63.15 & 75.73 & 85.60 & 97.01 \\
& LoRA                      & 72.51 & 90.86 & 50.59 & 62.31 & 74.98 & 85.62 & 96.99 \\
\midrule
\multirow{3}{*}{DINOv3}
& Frozen Encoder            & 71.02 & 90.97 & 50.20 & 62.93 & 75.17 & 80.83 & 92.04 \\
& Adapter                   & 73.36 & 91.93 & 51.29 & 63.28 & 76.89 & 84.68 & 95.74 \\
& LoRA                      & 73.81 & 92.05 & 52.25 & 64.07 & 76.54 & 85.32 & 95.58 \\
\bottomrule
\end{tabular}
\end{table*}

\begin{table*}[!htb]
\centering
\caption{Ablation results of the MFCE decoder in DINO3CD  (IoU \%).}
\label{tab:peftcd_dino3cd_ablation}
\begin{tabular}{l l c c c c c c c} % ← 9 列
\toprule
\textbf{PEFT} & \textbf{Decoder} & \textbf{SYSU-CD} & \textbf{WHUCD} & \textbf{S2Looking} & \textbf{MSRSCD} & \textbf{MLCD} & \textbf{LEVIR-CD} & \textbf{CDD} \\
\midrule
\multirow{2}{*}{Adapter} 
& Baseline & 70.66 & 90.96 & 49.42 & 61.83 & 75.42 & 83.69 & 94.87 \\
& MFCE     & 73.36 & 91.93 & 51.29 & 63.28 & 76.89 & 84.68 & 95.74 \\
\midrule
\multirow{2}{*}{LoRA}
& Baseline & 72.52 & 90.16 & 50.58 & 62.59 & 75.48 & 84.45 & 94.82 \\
& MFCE     & 73.81 & 92.05 & 52.25 & 64.07 & 76.54 & 85.32 & 95.58 \\
\bottomrule
\end{tabular}
\end{table*}

\subsection{Ablation Study}

As shown in Table~\ref{tab:peftcd_ablation}, PEFT delivers consistent gains over the Frozen Encoder baseline on both backbones (SAM2 and DINOv3), demonstrating that, with most parameters frozen, injecting only a small number of trainable modules can effectively adapt the model to the notion of “change.” Overall, DINOv3+LoRA attains the best or tied-best IoU on most datasets: it reaches 73.81/92.05/52.25/64.07 on SYSU-CD/WHUCD/S2Looking/MSRSCD, respectively, with substantial improvements over their frozen baselines. On MLCD, DINOv3+Adapter slightly outperforms LoRA (76.89 vs.\ 76.54), indicating that Adapter offers better robustness on datasets with more complex geomorphological changes and scale distributions. By contrast, the SAM2 backbone shows clearer advantages on LEVIR-CD and CDD: SAM2+LoRA achieves 85.62 on LEVIR-CD, while SAM2+Adapter yields the best IoU of 97.01 on CDD, reflecting the positive effect of large-scale segmentation priors for delineating building boundaries and suppressing seasonal appearance variations. In summary, DINOv3+LoRA provides consistent cross-scenario benefits attributable to DINOv3’s stronger generalization from pretraining on a larger and more diverse corpus than SAM2 whereas SAM2 exhibits finer boundary recognition in specific scenarios.

To further validate the effectiveness of the decoder design in the DINO3CD architecture, we conducted an additional set of ablation study focusing on the contribution of the proposed Multi-layer Fusion and Contextual Enhancement (MFCE) decoder, as reported in Table~\ref{tab:peftcd_dino3cd_ablation}. In this study, the Baseline decoder refers to a simplified design that refines and averages the feature maps from four DINOv3 layers, followed by a four-stage progressive upsampling. In contrast, the MFCE decoder incorporates the proposed multi-layer fusion and contextual enhancement mechanism. The results show that, regardless of whether Adapter or LoRA is adopted as the PEFT strategy, the MFCE decoder consistently outperforms the baseline decoder, which only upsamples the final feature representation. These findings confirm our hypothesis: for ViT-based backbones lacking an inherent feature pyramid, the MFCE decoder effectively integrates same-scale features across multiple layers and strengthens contextual awareness, thereby compensating for the limitations in spatial detail recovery and producing more precise and accurate change detection results.

\begin{table*}[!htb]
\centering
\caption{Models with trainable parameters and IoU performance across seven datasets.}
\label{tab:trainable_models_7datasets}
\resizebox{\textwidth}{!}{
\begin{tabular}{l c c c c c c c c}
\toprule
\textbf{Model} & \textbf{Trainable Params (M)} & \textbf{SYSU-CD} & \textbf{WHU-CD} & \textbf{S2Looking} & \textbf{MSRSCD} & \textbf{MLCD} & \textbf{CDD} & \textbf{LEVIR-CD} \\
\midrule
CDNeXt~\cite{wei_robust_2024} & 242.66 & 68.57 & 90.35 & 50.05 & 61.59 & --    & --    & --    \\
RSMamba~\cite{rs-mamba} &  51.59 & 68.94 & --    & --    & --    & --    & --    & 83.66 \\
ChangeMamba~\cite{chen2024changemamba} &  49.94 & 70.70 & --    & --    & --    & --    & --    & --    \\
HCGMNet~\cite{Han2023HCGMNetAH} &  47.32 & --    & 90.10 & --    & --    & --    & --    & 85.26 \\
ChangeFormer~\cite{bandara2022transformer} &  41.02 & 67.42 & --    & --    & 56.96 & --    & --    & 82.66 \\
SwinSUNet~\cite{swinsunet} &  39.28 & 68.42 & --    & --    & --    & --    & 88.65 & --    \\
CGNet~\cite{han_change_2023} &  38.99 & 66.55 & 90.41 & 47.41 & --    & --    & --    & 85.40 \\
ConvFormer-CD~\cite{Yang2025ConvFormerCDHC} &  37.72 & --    & --    & 51.51 & --    & --    & --    & 84.23 \\
ScratchFormer~\cite{Noman2023RemoteSC} &  37.00 & --    & --    & --    & --    & --    & 95.85 & 84.63 \\
TransUNetCD~\cite{transUnet} &  28.37 & 66.42 & --    & 54.41 & --    & --    & 94.50 & 83.67 \\
DCMamba~\cite{DC-mamba} &  17.35 & 72.72 & --    & --    & --    & --    & --    & --    \\
STANet~\cite{chen_spatial-temporal_2020} &  12.21 & 57.22 & --    & --    & 54.52 & 60.87 & --    & 81.85 \\
CDMamba~\cite{zhang_cdmamba_2025} &  11.90 & 71.64 & --    & --    & --    & --    & --    & --    \\
ELGCNet~\cite{m_noman_elgc-net_2024} &  10.57 & --    & --    & --    & --    & --    & 94.50 & 83.83 \\
ISDANet~\cite{h_ren_interactive_2025} &   6.86 & 70.90 & --    & --    & --    & 66.52 & --    & 83.63 \\
DMINet~\cite{feng_change_2023} &   6.76 & 69.60 & --    & 48.33 & 61.24 & 74.03 & --    & 82.99 \\
BASNet~\cite{z_wang_bitemporal_2024} &   4.19 & 69.71 & --    & --    & 61.31 & 70.55 & 93.29 & 83.11 \\
BIT~\cite{chen_remote_2022} &   3.01 & 66.03 & 88.22 & 47.94 & --    & 72.02 & --    & 80.68 \\
\midrule
\textbf{PeftCD (SAM2 Adapter)} & 11.00 & 71.91 & 91.81 & 49.97 & 63.15 & 75.73 & \textbf{97.01} & 85.60 \\
\textbf{PeftCD (SAM2 LoRA)} & 10.15 & 72.51 & 90.86 & 50.59 & 62.31 & 74.98 & 96.99 & \textbf{85.62} \\
\textbf{PeftCD (DINOv3 Adapter)} & 3.67 & 73.36 & 91.93 & 51.29 & 63.28 & \textbf{76.89} & 95.74 & 84.68 \\
\textbf{PeftCD (DINOv3 LoRA)} & \textbf{2.86} & \textbf{73.81} & \textbf{92.05} & \textbf{52.25} & \textbf{64.07} & 76.54 & 95.58 & 85.32 \\
\bottomrule
\end{tabular}}
\end{table*}

\section{Discussion}
\subsection{Decoding with Single-Scale ViT Backbones: Beyond Pyramids}
ViT backbones such as DINOv3 perform deep feature transformation at a fixed $1/16$ resolution, and thus inherently lack CNN/FPN-style multi-scale spatial pyramids. This limitation brings two direct issues: boundaries and small objects are prone to spatial aliasing and jagged artifacts during upsampling, and when relying solely on single-scale global semantics, local geometric and texture differences are often weakened even though they play a crucial role in change detection. To address these problems, we design a decoder that combines same-scale multi-layer deep attention fusion, ASPP-based contextual enhancement, and progressive upsampling, which bridges the semantic–geometric gap without altering the resolution of the backbone. Although same-scale multi-layer fusion largely alleviates the single-scale bottleneck, several limitations remain. The $1/16$ downsampling imposes an upper resolution bound for small objects, the fixed dilation rates in ASPP restrict adaptability to objects of different scales, and multi-layer fusion in large-scale models leads to high computational and memory cost. Future research can improve single-scale ViT decoders in two directions. On the one hand, more efficient multi-layer fusion modules or lightweight cross-layer interaction mechanisms can be developed to reduce overhead. On the other hand, auxiliary feature encoding branches can be incorporated to capture fine-grained image details, thereby enhancing boundary delineation and small-object recognition.

\subsection{PEFT Strategies for Change Detection: Adaptability and Limitations}
In this study, we systematically evaluate LoRA and Adapter as parameter-efficient fine-tuning (PEFT) strategies for change detection. Experimental results show that both methods achieve remarkable performance improvements while keeping the backbone frozen, especially on large-scale datasets such as WHUCD and CDD. This demonstrates that PEFT can quickly adapt VFMs to the downstream task of ``change'' without destroying their general capability. LoRA shows more significant gains when injected into attention projection layers, while Adapter exhibits stronger flexibility when inserted into different Transformer layers due to its bottleneck design. Nevertheless, PEFT still has limitations in change detection. When the change patterns differ substantially from those observed during pretraining, as in side-looking imagery from S2Looking, the adaptation capacity of PEFT becomes restricted and cannot fully compensate for domain gaps. In addition, because LoRA and Adapter are mostly injected independently into each layer, cross-layer interaction remains insufficient, which often leads to blurred boundaries when capturing fine-grained semantic changes. Finally, the current PEFT configurations are typically fixed, such as rank or bottleneck dimension, and lack mechanisms for adaptive adjustment according to dataset characteristics or task complexity.

\subsection{Parameter Efficiency of PeftCD}
A striking observation from Table~\ref{tab:peftcd_ablation} is the remarkable parameter efficiency of PeftCD. While existing models such as CDNeXt, RSMamba, and ChangeFormer rely on tens to hundreds of millions of trainable parameters, PeftCD adapts large vision backbones with only a few million or even fewer trainable weights. For example, PeftCD (DINOv3 LoRA) requires merely 2.86M parameters, yet it achieves the best or tied-best IoU scores across multiple benchmarks, including 73.81 on SYSU-CD, 92.05 on WHU-CD, 52.25 on S2Looking, and 64.07 on MSRSCD. These results not only surpass most existing architectures but also demonstrate consistent advantages over much larger backbones.  

The implications are twofold. First, parameter-efficient fine-tuning allows state-of-the-art performance without the prohibitive computational and memory costs of fully fine-tuned models, making PeftCD highly suitable for resource-constrained environments. Second, the effectiveness of lightweight modules such as LoRA and Adapter highlights that strong pretraining can be leveraged without sacrificing efficiency, as small task-specific adaptations are sufficient to capture complex change patterns. This balance of compactness and accuracy distinguishes PeftCD from conventional change detection networks and points to promising directions for future work on efficient model adaptation.

\section{Conclusion}
This paper proposes a novel change detection framework named PeftCD, which effectively addresses pseudo-change interference and generalization difficulties in remote sensing imagery by combining vision foundation models (e.g., SAM2, DINOv3) with parameter-efficient fine-tuning (PEFT) strategies. By freezing the majority of parameters in the backbone while only training a small number of additional modules, PeftCD successfully transfers the strong prior knowledge of VFMs to the change detection task. Experiments on seven public datasets, including SYSU-CD and WHUCD, demonstrate that PeftCD achieves state-of-the-art performance across multiple metrics, particularly excelling in delineating change boundaries and suppressing pseudo-changes. Overall, this study validates PeftCD as an efficient paradigm that balances accuracy, efficiency, and generalization, providing valuable insights for deploying large-scale foundation models in practical remote sensing applications.

\section{Acknowledgments}
The numerical calculations in this paper have been done on the supercomputing system in the Supercomputing Center of Wuhan University.

% \printbibliography
\bibliographystyle{IEEEtran}
\bibliography{PeftCD_ref}

\begin{thebibliography}{10}
\providecommand{\url}[1]{#1}
\csname url@samestyle\endcsname
\providecommand{\newblock}{\relax}
\providecommand{\bibinfo}[2]{#2}
\providecommand{\BIBentrySTDinterwordspacing}{\spaceskip=0pt\relax}
\providecommand{\BIBentryALTinterwordstretchfactor}{4}
\providecommand{\BIBentryALTinterwordspacing}{\spaceskip=\fontdimen2\font plus
\BIBentryALTinterwordstretchfactor\fontdimen3\font minus \fontdimen4\font\relax}
\providecommand{\BIBforeignlanguage}[2]{{%
\expandafter\ifx\csname l@#1\endcsname\relax
\typeout{** WARNING: IEEEtran.bst: No hyphenation pattern has been}%
\typeout{** loaded for the language `#1'. Using the pattern for}%
\typeout{** the default language instead.}%
\else
\language=\csname l@#1\endcsname
\fi
#2}}
\providecommand{\BIBdecl}{\relax}
\BIBdecl

\bibitem{Kirillov2023SegmentA}
\BIBentryALTinterwordspacing
A.~Kirillov, E.~Mintun, N.~Ravi, H.~Mao, C.~Rolland, L.~Gustafson, T.~Xiao, S.~Whitehead, A.~C. Berg, W.-Y. Lo, P.~Doll{\'a}r, and R.~B. Girshick, ``Segment anything,'' \emph{2023 IEEE/CVF International Conference on Computer Vision (ICCV)}, pp. 3992--4003, 2023. [Online]. Available: \url{https://api.semanticscholar.org/CorpusID:257952310}
\BIBentrySTDinterwordspacing

\bibitem{Caron2021EmergingPI}
\BIBentryALTinterwordspacing
M.~Caron, H.~Touvron, I.~Misra, H.~J'egou, J.~Mairal, P.~Bojanowski, and A.~Joulin, ``Emerging properties in self-supervised vision transformers,'' \emph{2021 IEEE/CVF International Conference on Computer Vision (ICCV)}, pp. 9630--9640, 2021. [Online]. Available: \url{https://api.semanticscholar.org/CorpusID:233444273}
\BIBentrySTDinterwordspacing

\bibitem{simeoni2025dinov3}
\BIBentryALTinterwordspacing
O.~Sim{\'e}oni, H.~V. Vo, M.~Seitzer, F.~Baldassarre, M.~Oquab, C.~Jose, V.~Khalidov, M.~Szafraniec, S.~Yi, M.~Ramamonjisoa, F.~Massa, D.~Haziza, L.~Wehrstedt, J.~Wang, T.~Darcet, T.~Moutakanni, L.~Sentana, C.~Roberts, A.~Vedaldi, J.~Tolan, J.~Brandt, C.~Couprie, J.~Mairal, H.~J{\'e}gou, P.~Labatut, and P.~Bojanowski, ``{DINOv3},'' 2025. [Online]. Available: \url{https://arxiv.org/abs/2508.10104}
\BIBentrySTDinterwordspacing

\bibitem{Radford2021LearningTV}
\BIBentryALTinterwordspacing
A.~Radford, J.~W. Kim, C.~Hallacy, A.~Ramesh, G.~Goh, S.~Agarwal, G.~Sastry, A.~Askell, P.~Mishkin, J.~Clark, G.~Krueger, and I.~Sutskever, ``Learning transferable visual models from natural language supervision,'' in \emph{International Conference on Machine Learning}, 2021. [Online]. Available: \url{https://api.semanticscholar.org/CorpusID:231591445}
\BIBentrySTDinterwordspacing

\bibitem{LORA}
\BIBentryALTinterwordspacing
E.~J. Hu, Y.~Shen, P.~Wallis, Z.~Allen-Zhu, Y.~Li, S.~Wang, L.~Wang, and W.~Chen, ``Lora: Low-rank adaptation of large language models,'' 2021. [Online]. Available: \url{https://arxiv.org/abs/2106.09685}
\BIBentrySTDinterwordspacing

\bibitem{adapter}
\BIBentryALTinterwordspacing
N.~Houlsby, A.~Giurgiu, S.~Jastrzebski, B.~Morrone, Q.~de~Laroussilhe, A.~Gesmundo, M.~Attariyan, and S.~Gelly, ``Parameter-efficient transfer learning for nlp,'' 2019. [Online]. Available: \url{https://arxiv.org/abs/1902.00751}
\BIBentrySTDinterwordspacing

\bibitem{ravi2024sam}
N.~Ravi, V.~Gabeur, Y.-T. Hu, R.~Hu, C.~Ryali, T.~Ma, H.~Khedr, R.~R{\"a}dle, C.~Rolland, L.~Gustafson \emph{et~al.}, ``Sam 2: Segment anything in images and videos,'' \emph{arXiv preprint arXiv:2408.00714}, 2024.

\bibitem{Zhou2021iBOTIB}
\BIBentryALTinterwordspacing
J.~Zhou, C.~Wei, H.~Wang, W.~Shen, C.~Xie, A.~L. Yuille, and T.~Kong, ``ibot: Image bert pre-training with online tokenizer,'' \emph{ArXiv}, vol. abs/2111.07832, 2021. [Online]. Available: \url{https://api.semanticscholar.org/CorpusID:244117494}
\BIBentrySTDinterwordspacing

\bibitem{sam_feature_extraction}
D.~Zhang, F.~Wang, L.~Ning, Z.~Zhao, J.~Gao, and X.~Li, ``Integrating sam with feature interaction for remote sensing change detection,'' \emph{IEEE Transactions on Geoscience and Remote Sensing}, vol.~62, pp. 1--11, 2024.

\bibitem{Ding2023AdaptingSA}
\BIBentryALTinterwordspacing
L.~Ding, K.~Zhu, D.~Peng, H.~Tang, K.~Yang, and L.~Bruzzone, ``Adapting segment anything model for change detection in vhr remote sensing images,'' \emph{IEEE Transactions on Geoscience and Remote Sensing}, vol.~62, pp. 1--11, 2023. [Online]. Available: \url{https://api.semanticscholar.org/CorpusID:261531371}
\BIBentrySTDinterwordspacing

\bibitem{self_sam}
S.~Saha and K.~Awadhiya, ``Integrating deep change vector analysis and sam for class-specific change detection,'' \emph{IEEE Journal of Selected Topics in Applied Earth Observations and Remote Sensing}, pp. 1--12, 2025.

\bibitem{Zhao2023AdaptingVT}
\BIBentryALTinterwordspacing
Y.~Zhao, Y.~Zhang, Y.~Dong, and B.~Du, ``Adapting vision transformer for efficient change detection,'' \emph{ArXiv}, vol. abs/2312.04869, 2023. [Online]. Available: \url{https://api.semanticscholar.org/CorpusID:266149789}
\BIBentrySTDinterwordspacing

\bibitem{Chen2024ChangeDA}
\BIBentryALTinterwordspacing
Y.~Chen, R.~Zhang, X.~Ning, H.~Zhang, Y.~He, Y.~Xie, and J.~Wang, ``Change dino: A unified transformer-based framework for object-level change detection and segmentation in remote sensing imagery,'' \emph{IGARSS 2024 - 2024 IEEE International Geoscience and Remote Sensing Symposium}, pp. 8585--8589, 2024. [Online]. Available: \url{https://api.semanticscholar.org/CorpusID:272432743}
\BIBentrySTDinterwordspacing

\bibitem{chen_spatial-temporal_2020}
\BIBentryALTinterwordspacing
H.~Chen and Z.~Shi, ``A spatial-temporal attention-based method and a new dataset for remote sensing image change detection,'' vol.~12, no.~10, p. 1662. [Online]. Available: \url{https://www.mdpi.com/2072-4292/12/10/1662}
\BIBentrySTDinterwordspacing

\bibitem{shi_deeply_2022}
\BIBentryALTinterwordspacing
Q.~Shi, M.~Liu, S.~Li, X.~Liu, F.~Wang, and L.~Zhang, ``A deeply supervised attention metric-based network and an open aerial image dataset for remote sensing change detection,'' vol.~60, pp. 1--16. [Online]. Available: \url{https://ieeexplore.ieee.org/document/9467555/}
\BIBentrySTDinterwordspacing

\bibitem{chen_remote_2022}
\BIBentryALTinterwordspacing
H.~Chen, Z.~Qi, and Z.~Shi, ``Remote sensing image change detection with transformers,'' vol.~60, pp. 1--14. [Online]. Available: \url{http://arxiv.org/abs/2103.00208}
\BIBentrySTDinterwordspacing

\bibitem{lin_transition_2023}
\BIBentryALTinterwordspacing
M.~Lin, G.~Yang, and H.~Zhang, ``Transition is a process: Pair-to-video change detection networks for very high resolution remote sensing images,'' vol.~32, pp. 57--71. [Online]. Available: \url{https://ieeexplore.ieee.org/document/9975266/}
\BIBentrySTDinterwordspacing

\bibitem{Xu2024HybridAT}
\BIBentryALTinterwordspacing
C.~Xu, Z.~Ye, L.~Mei, H.~Yu, J.~Liu, Y.~Yalikun, S.~Jin, S.~Liu, W.~Yang, and C.~Lei, ``Hybrid attention-aware transformer network collaborative multiscale feature alignment for building change detection,'' \emph{IEEE Transactions on Instrumentation and Measurement}, vol.~73, pp. 1--14, 2024. [Online]. Available: \url{https://api.semanticscholar.org/CorpusID:268375803}
\BIBentrySTDinterwordspacing

\bibitem{li_densely_2022}
\BIBentryALTinterwordspacing
Z.~Li, C.~Yan, Y.~Sun, and Q.~Xin, ``A densely attentive refinement network for change detection based on very-high-resolution bitemporal remote sensing images,'' vol.~60, pp. 1--18. [Online]. Available: \url{https://ieeexplore.ieee.org/document/9734050/}
\BIBentrySTDinterwordspacing

\bibitem{Jiang2022JointVL}
\BIBentryALTinterwordspacing
K.~Jiang, W.~Zhang, J.~Liu, F.~Liu, and L.~Xiao, ``Joint variation learning of fusion and difference features for change detection in remote sensing images,'' \emph{IEEE Transactions on Geoscience and Remote Sensing}, vol.~60, pp. 1--18, 2022. [Online]. Available: \url{https://api.semanticscholar.org/CorpusID:254343852}
\BIBentrySTDinterwordspacing

\bibitem{wei_robust_2024}
J.~Wei, K.~Sun, W.~Li, W.~Li, S.~Gao, S.~Miao, Q.~Zhou, and J.~Liu, ``Robust change detection for remote sensing images based on temporospatial interactive attention module,'' vol. 128, p. 103767.

\bibitem{z_wang_bitemporal_2024}
{Z. Wang}, {G. Gu}, {M. Xia}, {L. Weng}, and {K. Hu}, ``Bitemporal attention sharing network for remote sensing image change detection,'' vol.~17, pp. 10\,368--10\,379.

\bibitem{dong2024changeclip}
S.~Dong, L.~Wang, B.~Du, and X.~Meng, ``Changeclip: Remote sensing change detection with multimodal vision-language representation learning,'' \emph{ISPRS Journal of Photogrammetry and Remote Sensing}, vol. 208, pp. 53--69, 2024.

\bibitem{Chen2025AGFormerAA}
\BIBentryALTinterwordspacing
J.~Chen, D.~Wu, Q.~Ma, S.~Xu, and Y.~Zheng, ``Agformer: An anchor-guided transformer for class imbalance in remote sensing change detection,'' \emph{Pattern Recognit.}, vol. 168, p. 111839, 2025. [Online]. Available: \url{https://api.semanticscholar.org/CorpusID:278991383}
\BIBentrySTDinterwordspacing

\bibitem{zhao_exchanging_2023}
S.~Zhao, X.-l. Zhang, P.~Xiao, and G.~He, ``Exchanging dual-encoder–decoder: A new strategy for change detection with semantic guidance and spatial localization,'' vol.~61, pp. 1--16.

\bibitem{chen2024changemamba}
H.~Chen, J.~Song, C.~Han, J.~Xia, and N.~Yokoya, ``Changemamba: Remote sensing change detection with spatiotemporal state space model,'' \emph{IEEE Transactions on Geoscience and Remote Sensing}, vol.~62, pp. 1--20, 2024.

\bibitem{Li2025DifferenceEA}
\BIBentryALTinterwordspacing
H.~Li, S.~Cheng, and A.~Du, ``Difference enhancement and dependency-aware network for remote sensing images change detection,'' \emph{IEEE Journal of Selected Topics in Applied Earth Observations and Remote Sensing}, 2025. [Online]. Available: \url{https://api.semanticscholar.org/CorpusID:280678210}
\BIBentrySTDinterwordspacing

\bibitem{Zhan2025DifferenceAwareMF}
\BIBentryALTinterwordspacing
T.~Zhan, Q.~Tian, Y.~Zhu, J.~Lan, Q.~Dang, and M.~Gong, ``Difference-aware multiscale feature aggregation network for building change detection,'' \emph{IEEE Transactions on Geoscience and Remote Sensing}, vol.~63, pp. 1--15, 2025. [Online]. Available: \url{https://api.semanticscholar.org/CorpusID:277848794}
\BIBentrySTDinterwordspacing

\bibitem{li_overcoming_2025}
\BIBentryALTinterwordspacing
J.~Li, W.~He, Z.~Li, Y.~Guo, and H.~Zhang, ``Overcoming the uncertainty challenges in detecting building changes from remote sensing images,'' vol. 220, pp. 1--17. [Online]. Available: \url{https://www.sciencedirect.com/science/article/pii/S092427162400426X}
\BIBentrySTDinterwordspacing

\bibitem{f_zhou_dual-granularity_2025}
{F. Zhou}, {X. Zhang}, {H. Shuai}, {R. Hang}, {S. Zhu}, and {T. Geng}, ``Dual-granularity feature alignment for change detection in remote sensing images,'' vol.~18, pp. 4487--4497.

\bibitem{t_zhu_fta-net_2025}
{T. Zhu}, {Z. Zhao}, {M. Xia}, {J. Huang}, {L. Weng}, {K. Hu}, {H. Lin}, and {W. Zhao}, ``{FTA}-net: Frequency-temporal-aware network for remote sensing change detection,'' vol.~18, pp. 3448--3460.

\bibitem{Liu2025CWmambaLC}
\BIBentryALTinterwordspacing
Y.~Liu, G.~Cheng, Q.~Sun, C.~Tian, and L.~Wang, ``Cwmamba: Leveraging cnn-mamba fusion for enhanced change detection in remote sensing images,'' \emph{IEEE Geoscience and Remote Sensing Letters}, vol.~22, pp. 1--5, 2025. [Online]. Available: \url{https://api.semanticscholar.org/CorpusID:276819008}
\BIBentrySTDinterwordspacing

\bibitem{Li2025SAMMambaATC}
\BIBentryALTinterwordspacing
Y.~Li, W.~Liu, E.~Li, L.~Zhang, and X.~Li, ``Sam-mamba:a two-stage change detection network combining the adapting segment anything and mamba models,'' \emph{IEEE Journal of Selected Topics in Applied Earth Observations and Remote Sensing}, 2025. [Online]. Available: \url{https://api.semanticscholar.org/CorpusID:280764868}
\BIBentrySTDinterwordspacing

\bibitem{Zhou2025DepthCDDP}
\BIBentryALTinterwordspacing
N.~Zhou, M.~Zhou, and H.~Sui, ``Depthcd: Depth prompting in 2d remote sensing imagery change detection,'' \emph{ISPRS Journal of Photogrammetry and Remote Sensing}, 2025. [Online]. Available: \url{https://api.semanticscholar.org/CorpusID:279383174}
\BIBentrySTDinterwordspacing

\bibitem{Liu2025CDSTMambaTR}
\BIBentryALTinterwordspacing
S.~Liu, S.~Wang, W.~Zhang, T.~Zhang, M.~Xu, M.~Yasir, and S.~Wei, ``Cd-stmamba: Toward remote sensing image change detection with spatio-temporal interaction mamba model,'' \emph{IEEE Journal of Selected Topics in Applied Earth Observations and Remote Sensing}, vol.~18, pp. 10\,471--10\,485, 2025. [Online]. Available: \url{https://api.semanticscholar.org/CorpusID:277705639}
\BIBentrySTDinterwordspacing

\bibitem{Zhang2020ADS}
\BIBentryALTinterwordspacing
C.~Zhang, P.~Yue, D.~Tapete, L.~Jiang, B.~Shangguan, L.~Huang, and G.~Liu, ``A deeply supervised image fusion network for change detection in high resolution bi-temporal remote sensing images,'' \emph{Isprs Journal of Photogrammetry and Remote Sensing}, vol. 166, pp. 183--200, 2020. [Online]. Available: \url{https://api.semanticscholar.org/CorpusID:225504855}
\BIBentrySTDinterwordspacing

\bibitem{m_liu_cnn-transformer_2022}
{M. Liu}, {Z. Chai}, {H. Deng}, and {R. Liu}, ``A {CNN}-transformer network with multiscale context aggregation for fine-grained cropland change detection,'' vol.~15, pp. 4297--4306.

\bibitem{Han2023HCGMNetAH}
\BIBentryALTinterwordspacing
C.~Han, C.~Wu, and B.~Du, ``Hcgmnet: A hierarchical change guiding map network for change detection,'' \emph{IGARSS 2023 - 2023 IEEE International Geoscience and Remote Sensing Symposium}, pp. 5511--5514, 2023. [Online]. Available: \url{https://api.semanticscholar.org/CorpusID:257050289}
\BIBentrySTDinterwordspacing

\bibitem{han_change_2023}
C.~Han, C.~Wu, H.~Guo, M.~Hu, J.~Li, and H.~Chen, ``Change guiding network: Incorporating change prior to guide change detection in remote sensing imagery,'' vol.~\{\}, pp. 1--17.

\bibitem{dong_efficientcd_2024}
\BIBentryALTinterwordspacing
S.~Dong, Y.~Zhu, G.~Chen, and X.~Meng, ``Efficientcd: A new strategy for change detection based with bi-temporal layers exchanged,'' \emph{IEEE Transactions on Geoscience and Remote Sensing}, vol.~62, pp. 1--13, 2024. [Online]. Available: \url{https://api.semanticscholar.org/CorpusID:271334086}
\BIBentrySTDinterwordspacing

\bibitem{Fu2025BeyondCD}
\BIBentryALTinterwordspacing
S.~Fu, S.~Dong, and X.~Meng, ``Beyond cross-temporal difference: Style-aligned and fusion-difference learning for change detection,'' \emph{IEEE Transactions on Geoscience and Remote Sensing}, 2025. [Online]. Available: \url{https://api.semanticscholar.org/CorpusID:279703915}
\BIBentrySTDinterwordspacing

\bibitem{pei_feature_2022}
\BIBentryALTinterwordspacing
G.~Pei and L.~Zhang, ``Feature hierarchical differentiation for remote sensing image change detection,'' vol.~19, pp. 1--5. [Online]. Available: \url{https://ieeexplore.ieee.org/document/9837915/}
\BIBentrySTDinterwordspacing

\bibitem{ding2024adapting}
L.~Ding, K.~Zhu, D.~Peng, H.~Tang, K.~Yang, and L.~Bruzzone, ``Adapting segment anything model for change detection in vhr remote sensing images,'' \emph{IEEE Transactions on Geoscience and Remote Sensing}, vol.~62, pp. 1--11, 2024.

\bibitem{feng_change_2023}
Y.~Feng, J.~Jiang, H.~Xu, and J.~Zheng, ``Change detection on remote sensing images using dual-branch multilevel intertemporal network,'' vol.~61, pp. 1--15.

\bibitem{Ying2025AEGLNetAM}
\BIBentryALTinterwordspacing
Z.~Ying, Y.~Zhou, Y.~Zhai, H.~Zhu, H.~Zhang, P.~Coscia, A.~Genovese, F.~Scotti, V.~Piuri, and C.~L.~P. Chen, ``Aegl-net: Adaptive multiscale global–local feature fusion network for remote sensing change detection,'' \emph{IEEE Transactions on Geoscience and Remote Sensing}, vol.~63, pp. 1--19, 2025. [Online]. Available: \url{https://api.semanticscholar.org/CorpusID:279096874}
\BIBentrySTDinterwordspacing

\bibitem{Han2025HFIFNetHF}
\BIBentryALTinterwordspacing
M.~Han, T.~Xu, Q.~Liu, X.~Yang, J.~Wang, and J.~Kong, ``Hfifnet: Hierarchical feature interaction network with multiscale fusion for change detection,'' \emph{IEEE Journal of Selected Topics in Applied Earth Observations and Remote Sensing}, vol.~18, pp. 4318--4330, 2025. [Online]. Available: \url{https://api.semanticscholar.org/CorpusID:275486271}
\BIBentrySTDinterwordspacing

\bibitem{j_chen_sganet_2025}
{J. Chen}, {S. Dong}, and {X. Meng}, ``{SGANet}: A siamese geometry-aware network for remote sensing change detection,'' vol.~18, pp. 6232--6248.

\bibitem{fang_changer_2022}
S.~Fang, K.~Li, and Z.~Li, ``Changer: Feature interaction is what you need for change detection,'' vol.~61, pp. 1--11.

\bibitem{Yang2025ConvFormerCDHC}
\BIBentryALTinterwordspacing
F.~Yang, M.~Li, W.~Shu, A.~Qin, T.~Song, C.~Gao, and G.-S. Xia, ``Convformer-cd: Hybrid cnn–transformer with temporal attention for detecting changes in remote sensing imagery,'' \emph{IEEE Transactions on Geoscience and Remote Sensing}, vol.~63, pp. 1--15, 2025. [Online]. Available: \url{https://api.semanticscholar.org/CorpusID:276587808}
\BIBentrySTDinterwordspacing

\bibitem{He2015DeepRL}
\BIBentryALTinterwordspacing
K.~He, X.~Zhang, S.~Ren, and J.~Sun, ``Deep residual learning for image recognition,'' \emph{2016 IEEE Conference on Computer Vision and Pattern Recognition (CVPR)}, pp. 770--778, 2015. [Online]. Available: \url{https://api.semanticscholar.org/CorpusID:206594692}
\BIBentrySTDinterwordspacing

\bibitem{bandara2022transformer}
W.~G.~C. Bandara and V.~M. Patel, ``A transformer-based siamese network for change detection,'' in \emph{IGARSS 2022-2022 IEEE International Geoscience and Remote Sensing Symposium}.\hskip 1em plus 0.5em minus 0.4em\relax IEEE, 2022, pp. 207--210.

\bibitem{Chen2021FCCDNFC}
\BIBentryALTinterwordspacing
P.~Chen, D.~Hong, Z.~Chen, X.~S. Yang, B.~Li, and B.~Zhang, ``Fccdn: Feature constraint network for vhr image change detection,'' \emph{ArXiv}, vol. abs/2105.10860, 2021. [Online]. Available: \url{https://api.semanticscholar.org/CorpusID:235166250}
\BIBentrySTDinterwordspacing

\bibitem{Hang2024AANetAA}
\BIBentryALTinterwordspacing
R.~Hang, S.~Xu, P.~Yuan, and Q.~Liu, ``Aanet: An ambiguity-aware network for remote-sensing image change detection,'' \emph{IEEE Transactions on Geoscience and Remote Sensing}, vol.~62, pp. 1--11, 2024. [Online]. Available: \url{https://api.semanticscholar.org/CorpusID:268204920}
\BIBentrySTDinterwordspacing

\bibitem{Ma2024EATDerEA}
\BIBentryALTinterwordspacing
J.~Ma, J.~Duan, X.~Tang, X.~Zhang, and L.~Jiao, ``Eatder: Edge-assisted adaptive transformer detector for remote sensing change detection,'' \emph{IEEE Transactions on Geoscience and Remote Sensing}, vol.~62, pp. 1--15, 2024. [Online]. Available: \url{https://api.semanticscholar.org/CorpusID:266368343}
\BIBentrySTDinterwordspacing

\bibitem{Liu2025NetworkAD}
\BIBentryALTinterwordspacing
S.~Liu, D.~Zhao, Y.~Zhou, Y.~Tan, H.~He, Z.~Zhang, and L.~Tang, ``Network and dataset for multiscale remote sensing image change detection,'' \emph{IEEE Journal of Selected Topics in Applied Earth Observations and Remote Sensing}, vol.~18, pp. 2851--2866, 2025. [Online]. Available: \url{https://api.semanticscholar.org/CorpusID:275032911}
\BIBentrySTDinterwordspacing

\bibitem{Liu2025FullScaleCD}
\BIBentryALTinterwordspacing
------, ``Full-scale change detection network for remote sensing images based on deep feature fusion,'' \emph{IEEE Transactions on Geoscience and Remote Sensing}, vol.~63, pp. 1--13, 2025. [Online]. Available: \url{https://api.semanticscholar.org/CorpusID:277397556}
\BIBentrySTDinterwordspacing

\bibitem{h_ren_interactive_2025}
{H. Ren}, {M. Xia}, {L. Weng}, {H. Lin}, {J. Huang}, and {K. Hu}, ``Interactive and supervised dual-mode attention network for remote sensing image change detection,'' vol.~63, pp. 1--18.

\bibitem{Huang2025SAMBasedEF}
\BIBentryALTinterwordspacing
J.~Huang, J.~Bao, M.~Xia, and X.~Yuan, ``Sam-based efficient feature integration network for remote sensing change detection: A case study on macao sea reclamation,'' \emph{IEEE Journal of Selected Topics in Applied Earth Observations and Remote Sensing}, vol.~18, pp. 16\,916--16\,928, 2025. [Online]. Available: \url{https://api.semanticscholar.org/CorpusID:279733460}
\BIBentrySTDinterwordspacing

\bibitem{lin_feature_2017}
\BIBentryALTinterwordspacing
T.-Y. Lin, P.~Dollar, R.~Girshick, K.~He, B.~Hariharan, and S.~Belongie, ``Feature pyramid networks for object detection,'' in \emph{2017 {IEEE} Conference on Computer Vision and Pattern Recognition ({CVPR})}.\hskip 1em plus 0.5em minus 0.4em\relax {IEEE}, pp. 936--944. [Online]. Available: \url{http://ieeexplore.ieee.org/document/8099589/}
\BIBentrySTDinterwordspacing

\bibitem{lu_cross_2024}
\BIBentryALTinterwordspacing
K.~Lu, X.~Huang, R.~Xia, P.~Zhang, and J.~Shen, ``Cross attention is all you need: relational remote sensing change detection with transformer,'' \emph{GIScience \& Remote Sensing}, vol.~61, 2024. [Online]. Available: \url{https://api.semanticscholar.org/CorpusID:271326551}
\BIBentrySTDinterwordspacing

\bibitem{zhang_cdmamba_2025}
H.~Zhang, K.~Chen, C.~Liu, H.~Chen, Z.~Zou, and Z.~Shi, ``{CDMamba}: Incorporating local clues into mamba for remote sensing image binary change detection,'' vol.~63, pp. 1--16.

\bibitem{y_xing_sffce-cd_2025}
\BIBentryALTinterwordspacing
Y.~Xing, J.~Hu, B.~Jiang, Q.~Zhao, L.~Feng, and R.~Huang, ``Sffce-cd: Spatial and frequency feature cross enhancement for change detection,'' in \emph{IEEE International Conference on Acoustics, Speech, and Signal Processing}, 2025. [Online]. Available: \url{https://api.semanticscholar.org/CorpusID:276974126}
\BIBentrySTDinterwordspacing

\bibitem{m_noman_elgc-net_2024}
{M. Noman}, {M. Fiaz}, {H. Cholakkal}, {S. Khan}, and {F. S. Khan}, ``{ELGC}-net: Efficient local–global context aggregation for remote sensing change detection,'' vol.~62, pp. 1--11.

\bibitem{zhang_global-aware_2023}
R.~Zhang, H.~Zhang, X.~Ning, X.~Huang, J.~Wang, and W.~Cui, ``Global-aware siamese network for change detection on remote sensing images,'' vol. 199, pp. 61--72.

\bibitem{Zhang2025ADS}
\BIBentryALTinterwordspacing
Y.~Zhang, L.~Yang, L.~Zhu, C.~Zhou, Y.~A. Nanehkaran, and J.~Wang, ``A deep supervised change detection network based on context-rich information,'' \emph{IEEE Journal of Selected Topics in Applied Earth Observations and Remote Sensing}, 2025. [Online]. Available: \url{https://api.semanticscholar.org/CorpusID:280779334}
\BIBentrySTDinterwordspacing

\bibitem{d_sidekejiang_mldfnet_2025}
{D. Sidekejiang}, {P. Zheng}, and {L. Wang}, ``{MLDFNet}: A multilabel dual-flow network for change detection in bitemporal remote sensing images,'' vol.~18, pp. 4867--4880.

\bibitem{Noman2023RemoteSC}
\BIBentryALTinterwordspacing
M.~Noman, M.~Fiaz, H.~Cholakkal, S.~Narayan, R.~M. Anwer, S.~H. Khan, and F.~S. Khan, ``Remote sensing change detection with transformers trained from scratch,'' \emph{IEEE Transactions on Geoscience and Remote Sensing}, vol.~62, pp. 1--14, 2023. [Online]. Available: \url{https://api.semanticscholar.org/CorpusID:258108040}
\BIBentrySTDinterwordspacing

\bibitem{li_dual_2025}
\BIBentryALTinterwordspacing
Z.~Li, Z.~Zhang, M.~Li, L.~Zhang, X.~Peng, R.~He, and L.~Shi, ``Dual fine-grained network with frequency transformer for change detection on remote sensing images,'' vol. 136, p. 104393. [Online]. Available: \url{https://www.sciencedirect.com/science/article/pii/S1569843225000408}
\BIBentrySTDinterwordspacing

\bibitem{x_li_dsfi-cd_2025}
{X. Li}, {Y. Tan}, {K. Liu}, {X. Wang}, and {X. Zhou}, ``{DSFI}-{CD}: Diffusion-guided spatial-frequency-domain information interaction for remote sensing image change detection,'' vol.~63, pp. 1--18.

\bibitem{c_tao_hasnet_2025}
{C. Tao}, {D. Kuang}, {Z. Huang}, {C. Peng}, and {H. Li}, ``{HASNet}: A foreground association-driven siamese network with hard sample optimization for remote sensing image change detection,'' vol.~63, pp. 1--16.

\bibitem{Mao2022MFATNetMF}
\BIBentryALTinterwordspacing
Z.~Mao, X.-Y. Tong, Z.~Luo, and H.~Zhang, ``Mfatnet: Multi-scale feature aggregation via transformer for remote sensing image change detection,'' \emph{Remote. Sens.}, vol.~14, p. 5379, 2022. [Online]. Available: \url{https://api.semanticscholar.org/CorpusID:253220641}
\BIBentrySTDinterwordspacing

\bibitem{Liu2023AnAM}
\BIBentryALTinterwordspacing
W.~Liu, Y.-Y. Lin, W.~Liu, Y.~Yu, and J.~Li, ``An attention-based multiscale transformer network for remote sensing image change detection,'' \emph{ISPRS Journal of Photogrammetry and Remote Sensing}, 2023. [Online]. Available: \url{https://api.semanticscholar.org/CorpusID:259931492}
\BIBentrySTDinterwordspacing

\bibitem{Zhang2024B2CNetAP}
\BIBentryALTinterwordspacing
Z.~Zhang, L.~Bao, S.~Xiang, G.~Xie, and R.~Gao, ``B2cnet: A progressive change boundary-to-center refinement network for multitemporal remote sensing images change detection,'' \emph{IEEE Journal of Selected Topics in Applied Earth Observations and Remote Sensing}, vol.~17, pp. 11\,322--11\,338, 2024. [Online]. Available: \url{https://api.semanticscholar.org/CorpusID:270272635}
\BIBentrySTDinterwordspacing

\bibitem{Song2022RemoteSI}
\BIBentryALTinterwordspacing
X.~Song, Z.~Hua, and J.~Li, ``Remote sensing image change detection transformer network based on dual-feature mixed attention,'' \emph{IEEE Transactions on Geoscience and Remote Sensing}, vol.~60, pp. 1--16, 2022. [Online]. Available: \url{https://api.semanticscholar.org/CorpusID:252605199}
\BIBentrySTDinterwordspacing

\bibitem{Yuan2022STransUNetAS}
\BIBentryALTinterwordspacing
J.~Yuan, L.~Wang, and S.~Cheng, ``Stransunet: A siamese transunet-based remote sensing image change detection network,'' \emph{IEEE Journal of Selected Topics in Applied Earth Observations and Remote Sensing}, vol.~15, pp. 9241--9253, 2022. [Online]. Available: \url{https://api.semanticscholar.org/CorpusID:253333483}
\BIBentrySTDinterwordspacing

\bibitem{Huang2025LCCDMambaVS}
\BIBentryALTinterwordspacing
J.~Huang, X.~Yuan, C.~T. Lam, Y.~Wang, and M.~Xia, ``Lccdmamba: Visual state space model for land cover change detection of vhr remote sensing images,'' \emph{IEEE Journal of Selected Topics in Applied Earth Observations and Remote Sensing}, vol.~18, pp. 5765--5781, 2025. [Online]. Available: \url{https://api.semanticscholar.org/CorpusID:275776209}
\BIBentrySTDinterwordspacing

\bibitem{Ji2019FullyCN}
\BIBentryALTinterwordspacing
S.~Ji, S.~Wei, and M.~Lu, ``Fully convolutional networks for multisource building extraction from an open aerial and satellite imagery data set,'' \emph{IEEE Transactions on Geoscience and Remote Sensing}, vol.~57, pp. 574--586, 2019. [Online]. Available: \url{https://api.semanticscholar.org/CorpusID:57190346}
\BIBentrySTDinterwordspacing

\bibitem{Lebedev2018CHANGEDI}
\BIBentryALTinterwordspacing
M.~Lebedev, Y.~V. Vizilter, O.~Vygolov, V.~A. Knyaz, and A.~Y. Rubis, ``Change detection in remote sensing images using conditional adversarial networks,'' \emph{The International Archives of the Photogrammetry, Remote Sensing and Spatial Information Sciences}, 2018. [Online]. Available: \url{https://api.semanticscholar.org/CorpusID:57660599}
\BIBentrySTDinterwordspacing

\bibitem{Shen2021S2LookingAS}
\BIBentryALTinterwordspacing
L.~Shen, Y.~Lu, H.~Chen, H.~Wei, D.~Xie, J.~Yue, R.~Chen, Y.~Zhang, A.~Zhang, S.~Lv, and B.~Jiang, ``S2looking: A satellite side-looking dataset for building change detection,'' \emph{Remote. Sens.}, vol.~13, p. 5094, 2021. [Online]. Available: \url{https://api.semanticscholar.org/CorpusID:236134438}
\BIBentrySTDinterwordspacing

\bibitem{rs-mamba}
S.~Zhao, H.~Chen, X.~Zhang, P.~Xiao, L.~Bai, and W.~Ouyang, ``Rs-mamba for large remote sensing image dense prediction,'' \emph{IEEE Transactions on Geoscience and Remote Sensing}, vol.~62, pp. 1--14, 2024.

\bibitem{swinsunet}
C.~Zhang, L.~Wang, S.~Cheng, and Y.~Li, ``Swinsunet: Pure transformer network for remote sensing image change detection,'' \emph{IEEE Transactions on Geoscience and Remote Sensing}, vol.~60, pp. 1--13, 2022.

\bibitem{transUnet}
Q.~Li, R.~Zhong, X.~Du, and Y.~Du, ``Transunetcd: A hybrid transformer network for change detection in optical remote-sensing images,'' \emph{IEEE Transactions on Geoscience and Remote Sensing}, vol.~60, pp. 1--19, 2022.

\bibitem{DC-mamba}
\BIBentryALTinterwordspacing
J.~Zhang, R.~Chen, F.~Liu, H.~Liu, B.~Zheng, and C.~Hu, ``Dc-mamba: A novel network for enhanced remote sensing change detection in difficult cases,'' \emph{Remote Sensing}, vol.~16, no.~22, 2024. [Online]. Available: \url{https://www.mdpi.com/2072-4292/16/22/4186}
\BIBentrySTDinterwordspacing

\end{thebibliography}

% Biography text here without a photo.

% \begin{IEEEbiography}[{\includegraphics[width=1in,height=1.25in,clip,keepaspectratio]{fig1.png}}]{IEEE Publications Technology Team}
% In this paragraph you can place your educational, professional background and research and other interests.\end{IEEEbiography}

\end{document}